\documentclass[b4paper,10pt]{article}
\usepackage[utf8]{inputenc}
\usepackage[english]{babel}

\usepackage[margin={1.5cm,2.75cm}]{geometry}
\usepackage[font=small]{caption}
\usepackage{hyperref}
\usepackage{multirow}
\usepackage{url}
\usepackage{natbib}
\usepackage{amsmath,amsfonts,amssymb}
\usepackage{graphicx}
\usepackage{array}
\usepackage{booktabs}
\usepackage{multirow}
\usepackage{multicol}
\usepackage{float}

\let\oldtabular\tabular 
\renewcommand{\tabular}{\footnotesize\oldtabular}

\def\f{\mathbf{f}}
\def\x{\mathbf{x}}

\def\Gamma{\mathrm{Gamma}}
\def\k{\mathbf{k}}
\def\1{\mathbf{1}}
\def\R{\mathbb{R}}
\def\E{\mathbb{E}}

\def\k{\mathbf{k}}
\def\y{\mathbf{y}}

\def\bs{\boldsymbol\sigma}
\def\e{\varepsilon}
\def\0{\mathbf{0}}
\def\N{\mathcal{N}}
\def\GP{\mathcal{GP}}
\def\diag{\mathrm{diag}}
\def\DDG{\Delta\Delta\text{G}}
\def\DG{\Delta\text{G}}

\def\byE{\mathbf{y}_{E}}

\def\bsRT{\boldsymbol{\sigma}_{T}}
\def\bmu{\boldsymbol\mu}
\def\tr{\mathrm{tr}}
 
\newcommand{\cov}{\mathbf{cov}} 

\def\rmse{\mathrm{rmse}}

\begin{document}

\title{mGPfusion: Predicting protein stability changes with \\
Gaussian process kernel learning and data fusion}
\author{\large{Emmi Jokinen$^{1}$, Markus Heinonen$^{1,2}$ and Harri L\"ahdesm\"aki$^{1}$} \\ 
$^{1}$Department of Computer Science, Aalto University, 02150 Espoo, Finland\\ 
$^{2}$Helsinki Institute for Information Technology, Finland}
\date{\vspace{-0ex}}

\maketitle
\begin{center}
\begin{minipage}{0.95\textwidth}
\abstract{
\noindent\textbf{Motivation:} 
Proteins are commonly used by biochemical industry for numerous processes. Refining these proteins' properties via mutations causes stability effects as well. Accurate computational method to predict how mutations affect protein stability are necessary to facilitate efficient protein design. However, accuracy of predictive models is ultimately constrained by the limited availability of experimental data. \\ 
\textbf{Results:} 
We have developed mGPfusion, a novel Gaussian process (GP) method for predicting protein's stability changes upon single and multiple mutations. This method complements the limited experimental data with large amounts of molecular simulation data. We introduce a Bayesian data fusion model that re-calibrates the experimental and \emph{in silico} data sources and then learns a predictive GP model from the combined data. Our protein-specific model requires experimental data only regarding the protein of interest and performs well even with few experimental measurements. The mGPfusion models proteins by contact maps and infers the stability effects caused by mutations with a mixture of graph kernels. Our results show that mGPfusion outperforms state-of-the-art methods in predicting protein stability on a dataset of 15 different proteins and that incorporating molecular simulation data improves the model learning and prediction accuracy. \\
\textbf{Availability:} Software implementation and datasets are available at \url{github.com/emmijokinen/mgpfusion} \\
\textbf{Contact:} \href{emmi.jokinen@aalto.fi}{{emmi.jokinen@aalto.fi}}\\
\vspace{5ex}
}
\end{minipage}
\end{center}

\begin{multicols}{2}
\section{Introduction}

Proteins are used in various applications by pharmaceutical, food, fuel, and many other industries and their usage is growing steadily \citep{industrialenzymes,sanchez2010enzymes}. Proteins have important advantages over chemical catalysts, as they are derived from renewable resources, are biodegradable and are often highly selective \citep{directedenzymes}. Protein engineering is used to further improve the properties of proteins, for example to enhance their catalytic activity, modify their substrate specificity or to improve their thermostability \citep{rapley}. Increasing the stability is an important aspect of protein engineering, as the proteins used in industry should be stable in the industrial process conditions, which often involve higher than ambient temperature and non-aqueous solvents \citep{industry}. The properties of a protein are modified by introducing alterations to its amino acid sequence. Mutations in general tend to be destabilising, and if too many destabilising mutations are implemented, the protein may not remain functional without compensatory stabilising mutations \citep{tokuriki}.

The stability of a protein can be defined as the difference in Gibbs energy $\DG$ between the folded and unfolded (or native and denaturated) state of the protein. More precisely, the Gibbs energy difference determines the thermodynamic stability $\DG_\mathrm{t}$ of the protein, as it does not take into account the kinetic stability $\DG_\mathrm{k}$ which determines the energy needed for the transition between the folded and unfolded states \citep{mpoc} (see Supplementary Figure~S1).
Here we will consider only the thermodynamic stability and from now on it will be referred to merely as stability $\DG$.

The effect of mutations can be defined by the change they cause to the Gibbs energy $\DG$, denoted as $\DDG$ \citep{measuring}. To comprehend the significance of stability changes upon mutations, we can consider globular proteins, the most common type of enzymes, whose polypeptide chain is folded up in a compact ball-like shape with an irregular surface \citep{alberts}. These proteins are only marginally stable and the difference in Gibbs energy between the folded and unfolded state is only about 5--15 kcal/mol, which is not much more than the energy of a single hydrogen bond that is about 2--5 kcal/mol \citep{proteinStructure}. Therefore, even one mutation that breaks a hydrogen bond can prevent a protein from folding properly.
 
The protein stability can be measured with many techniques, including thermal, urea and guanidinium chloride (GdmCl) denaturation curves that are determined as the fraction of unfolded proteins at different temperatures or at different concentrations of urea or GdmCl \citep{lem}. Some of the experimentally measured stability changes upon mutations have been gathered in thermodynamic databases such as Protherm \citep{protherm}.

A variety of computational methods have been introduced to predict the stability changes upon mutations more effortlessly than through experimental measurements. These methods utilise physics or knowledge-based potentials \citep{rosetta3}, their combinations, or different machine learning methods. The machine learning methods utilise support vector machines (SVM) \citep{imutant2,threestate,istable,mupro,folkman,gradingaaprop15,duet}, random forests \citep{prethermut,promaya}, neural networks \citep{pop2,neemo}, and Gaussian processes \citep{mcsm}. However, it has been assessed that although on average many of these methods provide good results, they tend to fail on details \citep{potapov}. In addition, many of these methods are able to predict the stability effects only for single-point mutations.

We introduce mGPfusion (mutation Gaussian Processes with data fusion), a method for predicting stability effects of both point and multiple mutations. mGPfusion is a protein-specific model -- in contrast to earlier stability predictors that aim to estimate arbitrary protein structure or sequence stabilities -- and achieves markedly higher accuracy while utilising data only from a single protein at a time. In contrast to earlier works that only use experimental data to train their models, we also combine exhaustive Rosetta \citep{rosetta3} simulated point mutation \emph{in silico} stabilities to our training data. 

A key part of mGPfusion is the automatic scaling of simulated data to better match the experimental data distribution based on those variants that have both experimental and simulated stability values. Furthermore, we estimate a variance resulting from the scaling, which places a higher uncertainty on very destabilising simulations. Our Gaussian process model then utilises the joint dataset with their estimated heteroscedastic variances and uses a mixture of graph kernels to assess the stability effects caused by changes in amino acid sequence according to 21 substitution models. Our experiments on a novel 15 protein dataset show a state-of-the-art stability prediction performance, which is also sustained when there is access only to a very few experimental stability measurements.

\begin{figure*}[tb]
\centerline{\includegraphics[width=\linewidth]{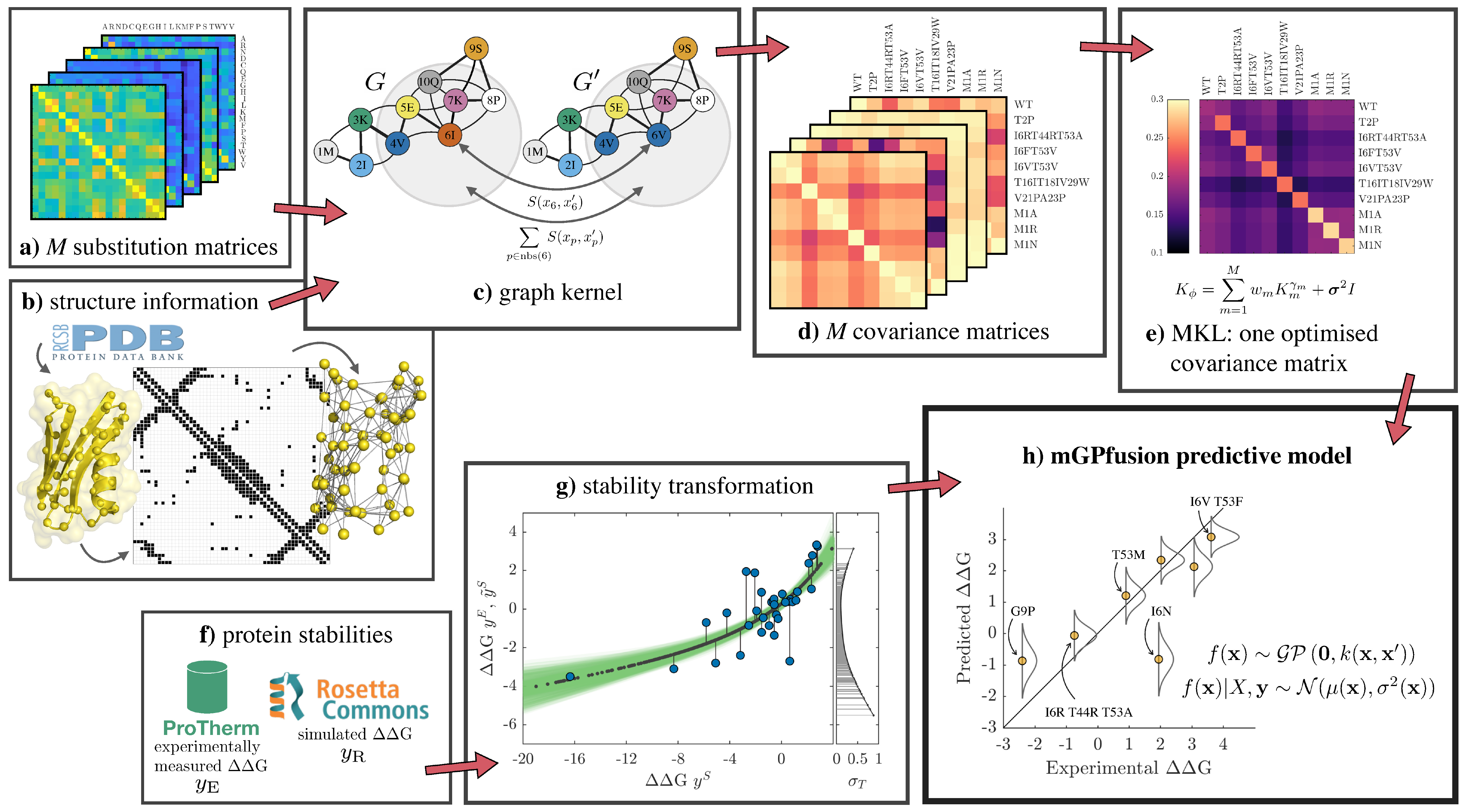}}
\caption{Pipeline illustration for mGPfusion. 
\textbf{a)} $M=21$ substitution matrices utilise different information sources and give scores to pairwise amino acid substitutions.
\textbf{b)} The wild-type structures from Protein Data Bank are modelled as contact graphs. 
\textbf{c)} The graph kernel measures similarity of two sequences by a substitution model $S$ over all positions $p$ and their neighbourhoods $\mathrm{nbs}(p)$ in the contact graph.
\textbf{d)} Each substitution matrix is used to create a separate covariance matrix.
\textbf{e)} Multiple kernel learning (MKL) is used for finding the optimal combination of the base kernels. The kernel matrix measures variant similarities.
\textbf{f)} Experimentally measured $\DDG$ values $y^E$ are gathered from Protherm and Rosetta's ddg monomer application is used to simulate the stability effects $y^S$ for all single point mutations. 
\textbf{g)} Bayesian scaling for the simulated values $y^S$ at the x-axis. Possible scalings are coloured with green and the chosen scaling from $y^S$ into scaled values $\tilde{y}^S$ is marked by black dots. The scaling is fitted to a subset of experimentally measured stabilities $y^E$ (circles). 
\textbf{h)} The stability predictive GP model is trained using experimental and simulated data through the kernel matrix.}\label{fig:pipeline}
\end{figure*}

\section{Methods}

Following \citet{mcsm} we choose a Bayesian model family of Gaussian processes for prediction of mutation effects on protein stability due to its inherent ability to handle uncertainty in a principle way. Bayesian modelling is a natural approach for combining the experimental and simulated data distribution, while it is also suitable for learning the underlying mixture of substitution models that governs the mutational process.

The pipeline for mGPfusion is presented in Figure~\ref{fig:pipeline}. The first part of mGPfusion consists of collection of \emph{in silico} and experimental datasets discussed in Section~\ref{sec:data}, the scaling of the \emph{in silico} dataset in Section~\ref{sec:scaling} and the fusion of these two datasets in Section~\ref{sec:fusion}. The second part consists of the Gaussian process model described in Section~\ref{sec:gp} with detailed description of the graph kernels in Sections~\ref{sec:graph}-\ref{sec:mkl} and model inference in Section~\ref{sec:inference}. Finally, the evaluation criteria used are described in Section~\ref{sec:eval}.

\subsection{Experimental and \emph{in silico} data} \label{sec:data}

Protherm is a database of numerical thermodynamic parameters for proteins and their mutants \citep{protherm}. From Protherm we gathered all proteins with at least 50 unique mutations whose $\DDG$ has been measured by thermal denaturation, and where a PDB code for a 3D structure of the protein was reported. We required the proteins to have at least 50 unique mutations, so that we would have a representative test set and get sufficiently reliable estimates of prediction accuracy on individual proteins and examine how the amount of experimental training data affects the accuracy of the model. The 3D structures are necessary for obtaining the connections between residues. We collected the 3D structures with the reported PDB codes from the Protein Databank, www.rcsb.org \citep{pdb2000}. The 15 proteins that fulfilled these requirements are listed in Table~\ref{tab:thermal}. We averaged the stability values for proteins with multiple measurements and ignored mutations to residues not present in their 3D structures. These data sets are available at \url{github.com/emmijokinen/mgpfusion}.
\begin{table}[H]
\resizebox{\linewidth}{!}{
\begin{tabular}{@{}l @{}ccccc@{}}
\toprule 
\multirow{2}{*}{Protein (organism)} & \multirow{2}{*}{PDB} & \multicolumn{3}{c}{mutations} \\
 & & all & \!\!point\!\! & point (sim) \\\midrule
T4 Lysozyme (\it{Enterobacteria phage T4})               & 2LZM & 349 & 264 & 3116\\
Barnase (\it{Bacillus amyloliquefaciens})                & 1BNI & 182 & 163 & 2052\\
Gene V protein (\it{Escherichia virus m13})              & 1VQB & 124 & 92  & 1634\\
Glycosyltransferase A (\it{Homo sapiens})                & 1LZI & 116 & 114 & 2470\\
Chymotrypsin inhibitor 2 (\it{Hordeum vulgare})          & 2CI2 & 98  & 77  & 1235\\
Protein G (\it{Streptococcus sp. gx7805})                & 1PGA & 89  & 34  & 1064\\
Ribonuclease H (\it{Escheria coli})                      & 2RN2 & 83  & 65  & 2945\\
Cold shock protein B (\it{Bacillus subtilis})            & 1CSP & 80  & 50  & 1273\\
Apomyoglobin (\it{Physeter catodon})                     & 1BVC & 80  & 56  & 2907\\
Hen egg white lysozyme (\it{Gallus gallus})              & 4LYZ & 63  & 50  & 2451\\
Ribonuclease A (\it{Bos taurus})                         & 1RTB & 57  & 50  & 2356\\
Peptidyl-prolyl cis-trans isomerase (\it{Homo sapiens})  & 1PIN & 56  & 56  & 2907\\
Ribonuclease T1 isozyme (\it{Aspergillus oryzae})        & 1RN1 & 53  & 48  & 1957\\
Ribonuclease (\it{Streptomyces auerofaciens})            & 1RGG & 54  & 45  & 1824\\
Bovine pancreatic trypsin inhibitor (\it{Bos taurus})    & 1BPI & 53  & 47  & 1102\\\midrule
total                                                    &      & 1537& 1211& 31293\\\bottomrule
\end{tabular}}
\caption{The 15 protein data from ProTherm database with counts of point mutations, all mutations, and of simulated point mutation stability changes.}
\label{tab:thermal}
\end{table}

We also generate simulated data of the stability effects of all possible single mutations of the proteins. Our method can utilise any simulated stability values. We used the ``ddG monomer'' application of Rosetta 3.6 \citep{rosetta3} using the high-resolution backrub-based protocol 16 recommended in \citet{kellogg}. The predictions $\y^S$ made with Rosetta are given in Rosetta Energy Units (REU). 
\citet{kellogg} suggest transformation $0.57 y^S$ for converting the predictions into physical units. The simulated data scaled this way is not as accurate as the experimental data, the correlation and root mean square error ($\rmse$) with respect to the experimental data are shown for all proteins in Table~\ref{tab:comparison} and for individual proteins in Supplementary Table~S2, on rows labelled Rosetta. For this reason, we use instead a Bayesian scaling described in the next section and different noise models for the experimental and simulated data, described in Section~\ref{sec:fusion}.

For each of the $15$ proteins, let $\x_i = (x_{i1}, \ldots, x_{iM})$ denote its $M$-length variant $i$ with positions $p$ labelled with residues $x_{ip} \in \{A,R,N,\ldots,V\}$. We denote the wild-type protein as $\x_0$. We collect 15 separate sets of simulated and experimental data. We denote the $N_E$ experimental variants of each protein as $X_E = (\x^E_1, \ldots, \x^E_{N_E})^T$ with the corresponding experimental stability values $\y_E= (y^E_1, \ldots, y^E_{N_E})^T \in \R^{N_E}$. Similarly, we denote the $N_S$ simulated observations as $X_S = (\x_1^S, \ldots, \x^S_{N_S})^T$ and $\y_S = (y_1^S, \ldots, y^S_{N_S})^T \in \R^{N_S}$.

\subsection{Bayesian scaling of \emph{in silico} data} \label{sec:scaling}

The described transformation from REU to physical units may not be optimal for all proteins. We therefore applied instead a linear-exponential scaling function to obtain \emph{scaled} Rosetta simulated stabilities $\tilde{y}^S$,
\begin{align}
\tilde{y}^S = g(y^S \,|\, \theta_j) = a_j e^{c_j y^S} + b_j y^S + d_j. \label{g} 
\end{align}
This scaling transforms the Rosetta simulations $y^S$ for each protein $j = 1, \ldots, 15$ to correspond better to the experimental data. The parameters $\theta_j = (a_j,b_j,c_j,d_j)$ define the weight $a_j$ and steepness $c_j$ of the exponential term, while the linear term has slope $b_j$ and intercept $d_j$. To avoid overfitting, we perform Bayesian linear regression and start by defining parameter prior $p(\theta_j) = p(a_j)p(b_j)p(c_j)p(d_j)$ that reflects our beliefs about realistic scalings having only moderate steepness: 
\begin{align}
\begin{split}
p(a_j) &= \mathrm{Gamma}(a_j \,|\, \alpha_a, \beta_a) \\ 
p(b_j) &= \mathrm{Beta}(1/2 \cdot b_j \,|\, \alpha_b, \beta_b) \\ 
p(c_j) &= \mathrm{Beta}(10/3 \cdot c_j \,|\, \alpha_c, \beta_c) \\
p(d_j) &= \N(d_j \,|\, \mu_d, \sigma_d^2). \label{eq:tprior}
\end{split}
\end{align}
The empirically selected hyperparameter values are listed in Supplementary Table~S1 and the priors are illustrated in Figure~S2.

We compute the posterior for $\theta_j$ using the subset of simulated data that have corresponding experimentally measured data:
$$p(\theta_j | \y_E, \y_S) \propto \prod_{i : \x_i \in X_E \cap X_S} \N \left(y^E_i \,|\, g(y^S_i | \theta_j), \sigma_n^2\right) p(\theta_j).$$
The product iterates over all $N_{E \cap S}$ simulated $\DDG$'s that have a matching experimentally observed value. The $\sigma_n^2$ is the scaling error variance, which was set to $\sigma_n^2 = 0.5$. The parameters $\theta$ for each protein were sampled using a random walk Metropolis-Hastings MCMC algorithm (the \texttt{mhsample} function in Matlab) for $N_{MC} = 10000$ samples with a burn-in set to $500$. The proposal distribution was selected to be a symmetric uniform distribution such that $[a_{s+1}, b_{s+1}, c_{s+1}, d_{s+1}] \sim U(a_s\pm0.4, b_s\pm 0.04, c_s\pm 0.04, d_s\pm0.4)$. Given the sample of scaling parameters $(\theta_j^{(s)})_{s=1}^{N_{MC}}$, we define the scaled simulated data as the average scaling over the MCMC sample, and record also the sample scaling variance
\begin{align}
\tilde{y}^S_i  &= \frac{1}{N_{MC}} \sum_{s=1}^{N_{MC}}  g(y^S_i | \theta_j^{(s)}) \\
\sigma_{T}^2(i) &= \frac{1}{N_{MC}} \sum_{s=1}^{N_{MC}} \left( g(y^S_i | \theta_j^{(s)}) - \tilde{y}^S_i \right)^2. \label{t-variance}
\end{align}
See Figure~\ref{fig:pipeline}~g) for an illustration of the scaling. We collect the scaled simulated value and its variance from each simulated point into vectors $\tilde{\y}_S = (\tilde{y}_1^S, \ldots, \tilde{y}_{N_S}^S)$ and  $\boldsymbol{\sigma}_T^2 = (\sigma_T^2(1), \ldots, \sigma_T^2(N_S)) \in \R^{N_S}$.

\subsection{Data fusion and noise models} \label{sec:fusion}

For each protein $j$, we organise its experimental data $(X_E,\byE)$ and transformed simulated data $(X_S,\tilde{\y}_S)$ along with the wild-type information $(\x_0,y_0)$ into a single joint dataset of variants $X = (\x_0, X_E, X_S)$ and stabilities $\y = (y_0, \y_E, \tilde{\y}_S)$ of size $\R^N$ where $N = 1 + N_E + N_S$ is the total number of simulated and experimental data points, including the wild-type. We assume heteroscedastic additive noise models for the three information sources
\begin{align}
y_0 &= f\left(\x_0\right) + \e_0,  & \e_0  &\sim \N\left(0, \sigma_0^2\right) \notag \\
y^E_i &= f\left(\x^E_i\right) + \e^E_i,  & \e_i^E  &\sim \N\left(0, \sigma_E^2\right)  \label{eq:noise} \\
\tilde{y}^S_i &= f\left(\x^S_i\right) + \e^S_i,  & \e_i^S  &\sim \N\left(0, \left(\sigma_E + \sigma_S + t \sigma_T(i) \right)^2 \right), \notag
\end{align}
where the observed values are noisy versions of the underlying `true' stability function $f(\x)$ corrupted by zero-mean noise with data source specific variances. We learn a Gaussian process based stability function $f(\x)$ in the next Section.

The Equations \eqref{eq:noise} encode that the experimental data are corrupted by a global experimental noise variance $\sigma_E^2$. The simulated stabilities are additionally corrupted by a global Rosetta simulator error variance $\sigma_S^2$, and by the value-dependent transformation variance $t \sigma_T^2(i)$ scaled by parameter $t$. The model then encapsulates that we trust the Rosetta data less than the experimental data. By definition, the $\DDG$ of the wild-type is zero ($y_0 = 0$) with very small assumed error, $\sigma_0 = 10^{-6}$. Note that $\bs_T^2$ are fixed by equation~\eqref{t-variance}, while we infer the optimal values for the remaining three free parameters $(\sigma_E, \sigma_R, t)$ (See Section \ref{sec:gp}). The parameters $\sigma_E^2$ and $\sigma_S^2$ are assigned priors
\begin{align}
\begin{split}
\sigma_E &\sim \Gamma(\sigma_E | \alpha_E, \beta_E) \\
\sigma_S &\sim \Gamma(\sigma_R | \alpha_S, \beta_S) \label{eq:sigmaprior}.
\end{split}
\end{align}
The values of these hyperparameters are shown in Supplementary Table~S1.

\subsection{Gaussian processes} \label{sec:gp}

We use a Gaussian process (GP) function $f$ to predict the stability $f(\x) \in \R$ of a protein variant $\x$. Gaussian processes are a family of non-parametric, non-linear Bayesian models~\citep{rasmussen}. A zero-mean GP prior 
\begin{align*}
f(\x) \sim \GP\left(0, k(\x,\x')\right),
\end{align*}
defines a distribution over functions $f(\x)$ whose mean and covariance are
\begin{align*}
\E[f(\x)] &= 0 \\
\cov[ f(\x),f(\x')] &= k(\x,\x').
\end{align*}
For any collection of protein variants $X = \x_1, \ldots, \x_N$, the function values follow a multivariate normal distribution $\f \sim \N(\0, K_{XX})$, where $\f = (f(\x_1), \ldots, f(\x_N))^T \in \R^N$, and where $K_{XX} \in \R^{N \times N}$ with $[K_{XX}]_{ij} = k(\x_i, \x_j)$. The key property of Gaussian processes is that they encode functions that predict similar stability values $f(\x),f(\x')$ for protein variants $\x,\x'$ that are similar, as encoded by the kernel $k(\x,\x')$. The key part of GP modelling is then to infer a kernel that measures the mutation's effects to the stability.

Let a dataset of noisy stability values from two sources be $\y \in \R^N$, the corresponding protein structures $X = (\x_i)_{i=1}^N$, and a new protein variant $\x_*$ whose stability we wish to predict. A Gaussian process defines a joint distribution over the observed values $\y$ of variants $X$, and the unknown function value $f(\x_*)$ of the unseen variant $\x_*$,
\begin{align*}
\begin{bmatrix} \y \\ f(\x_*) \end{bmatrix} 
\sim
\N\left(\0, \begin{bmatrix}
K_{XX} + \diag(\bs^2) & \k_{X*} \\
\k_{*X} & k(\x_*,\x_*)
\end{bmatrix}\right),
\end{align*}
where $\k_{X*} = \k_{*X}^T \in \R^{N}$ is a kernel vector with elements $k(\x_i, \x_*)$ for all $i=1,\ldots,N$, and where $\bs^2 = (\sigma_0^2, \sigma_E^2 \1^T, (\sigma_E \1^T + \sigma_S \1^T + t {\bs_T}^T )^2)^T$ collects final variances of the data points from equations~\eqref{eq:noise}. Here the exponents are elementwise. The conditional distribution gives the posterior distribution of the stability prediction as
\begin{align*}
f(\x_*) | (X,\y) \sim \N\left(\mu(\x_*), \sigma^2(\x_*)\right), 
\end{align*}
where the prediction mean and variance are
\begin{align*}
\mu(\x_*)    &= \k_{*X} \left(K_{XX} + \diag(\bs^2) \right)^{-1} \y,\\
\sigma^2(\x_*) &= k(\x_*,\x_*) - \k_{*X} \left(K_{XX} + \diag(\bs^2)\right)^{-1} \k_{X*}.
\end{align*}
Hence, in GP regression the stability predictions $\mu(\x_*) \pm \sigma(\x_*)$ will come with uncertainty estimates.

\subsection{Graph kernel}\label{sec:graph}

Next, we consider how to compute the similarity function $k(\x,\x')$ between two variants of the same protein structure. We will encode the 3D structural information of the two protein variants as a contact map and measure their similarity by the formalism of graph kernels \citep{vishwanathan2010}. 

We consider two residues to be in contact if their closest atoms are within 5 {\AA} of each other in the PDB structure, which is illustrated in Figure~\ref{fig:pipeline}~b). 
All variants of the same protein have the same length, with only different residues at mutating positions. Furthermore, we assume that all variants share the wild-type protein contact map.

To compare protein variants, we construct a weighted decomposition kernel (WDK) \citep{wdk} between two protein variants $\x = (x_1, \ldots, x_M)$ and $\x' = (x_1', \ldots, x_M')$ of length $M$,
\begin{align}
k(\x,\x') = \sum_{p=1}^M \left( S(x_p,x_p') \sum_{l \in \mathrm{nbs}(p)} S(x_l,x_l') \right)\hspace{-1mm},
\label{Ksingle}
\end{align}
where $\mathrm{nbs}(p)$ defines the set of neighbouring positions to position $p$, and $S$ is a substitution matrix. The kernel iterates over all positions $p$ and compares for each of them their residues through a substitution matrix $S(x_p,x_p')$. Furthermore, the similarity of the residues at each position is multiplied by the average similarity of the residues at its neighbouring positions $S(x_l,x_l')$. Hence, the kernel defines the similarity of two protein variants as the average position and neighbourhood similarity over all positions. The kernel matrix is normalised so that for two data points, the normalised kernel is $\hat{k}(x_p,x_p')=k(x_p,x_p')/\sqrt{k(x_p,x_p)k(x_p',x_p')}$, as defined by \citet{shawetaylor04}. The kernel is illustrated in Figure~\ref{fig:pipeline}~c).

The above WDK kernel allows us to compare the effects of multiple simultaneous mutations. However, as the wild type protein structure is used for all of the protein variants, changes that the mutations may cause to the protein structure are not taken into consideration. This may cause problems if mutations that alter the protein structure significantly are introduced -- especially if many of them are introduced simultaneously. On the other hand, substitution matrices that have their basis in sequence comparisons, should take these effects into account to some extend as these kinds of mutations are usually highly destabilising and do not occur often in nature. In the next section, we will discuss how we utilise different substitution matrices with multiple kernel learning.

\subsection{Substitution matrices and multiple kernel learning}\label{sec:mkl}

The BLOSUM substitution models have been a common choice for protein models \citep{giguere13}, while mixtures of substitution models were proposed by \citet{cichonska17}. BLOSUM matrices score amino acid substitutions by their appearances throughout evolution, as they compare the frequencies of different mutations in similar blocks of sequences \citep{henikoff1992}. However, there are also different ways to score amino acids substitutions, such as chemical similarity and neighbourhood selectivity \citep{tomii}. When the stability effects of mutations are evaluated, the frequency of an amino acid substitution in nature may not be the most important factor.

\begin{figure*}[tb]
\centerline{\includegraphics[width=\linewidth]{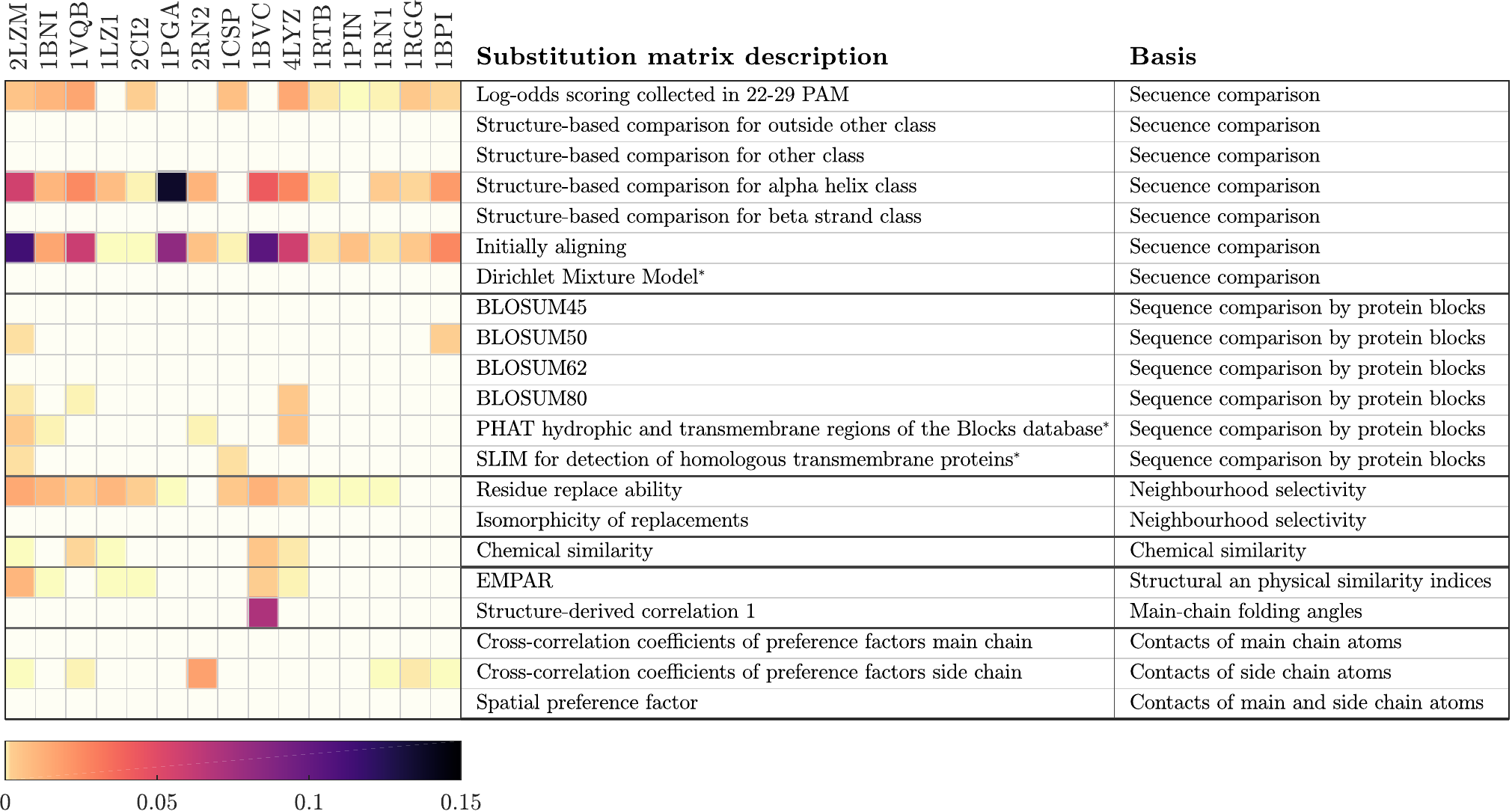}}
\caption{Average weights for kernels utilising the described substitution matrices from AAindex2, when GP models were trained with mutation level cross-validation. Basis for the substitution matrices are obtained from \citep{tomii}. $^*$ were added to AAindex2 in a later release, and their basis were not determined by \citet{tomii}.} \label{fig:kernelweights}
\end{figure*}

To take into account different measures of similarity between amino acids, we employed a set of 21 amino acid substitution matrices gathered from AAindex2\footnote{\texttt{http://www.genome.jp/aaindex/}} \citep{aaindex}. AAindex2 currently contains 94 substitution matrices. From these we selected those that had no gaps concerning substitutions between the 20 naturally occurring amino acids and scaled them between zero and one as
\begin{align}
S = \frac{\mathrm{S_0} - \min(\mathrm{S_0})+1}{\max(\mathrm{S_0}) - \min(\mathrm{S_0})+1}.
\end{align}
Out of these matrices, we only chose those 23 matrices that were positive semidefinite. Furthermore, there were two pairs of matrices that were extremely similar, and we only selected one matrix from each pair, ending up with 21 substitution matrices. These substitution matrices are used together with Equation~\ref{Ksingle} for computing 21 base kernel matrices. Finally, MKL is used to find an optimal combination of the base kernels of form
\begin{align}
    K_\phi =\sum_{m=1}^{21} w_m K_m^{(\gamma_m)},
\end{align}
where $w_m$ is a kernel specific weight, $\gamma_m$ is an (elementwise) exponent.
The elementwise exponent retains the SDP property of $K_\phi$ \citep{shawetaylor04}. We observe empirically that the optimal kernel weights $w_m$ tend to be sparse (See Figure~\ref{fig:kernelweights}).

The selected substitution matrices are listed in Figure~\ref{fig:kernelweights}. These matrices have different basis and through multiple kernel learning (MKL) our model learns which of these are important for inferring the stability effects that mutations cause on different proteins. The figure illustrates this by showing the average weights of the base kernel matrices obtained via the multiple kernel learning.

\subsection{Parameter inference} \label{sec:inference}

The complete model has five parameters $\phi = (\sigma_E, \sigma_S, t, \mathbf{w}, \boldsymbol{\gamma})$ to infer, of which the variance parameters $(\sigma_E, \sigma_S, t)$ parameterise the joint data variance $\bs_\phi^2$, while the MKL parameters $\mathbf{w}=(w_1,\ldots,w_{21})$ and $\boldsymbol{\gamma}=(\gamma_1,\ldots,\gamma_{21})$ parameterise the kernel matrix $K_\phi$. 
In a Gaussian process model these can be jointly optimised by the marginal (log) likelihood with priors
\begin{align}
\begin{split}
\log p(\y | \phi) &p(\sigma_E) p(\sigma_R) = \log \int p(\y | \f, \phi) p(\f | \phi) p(\sigma_E) p(\sigma_R) d\f \\
  &\hspace{-10mm} \propto - \frac{1}{2} \y^T (K_\phi + \diag(\bs_\phi^2))^{-1}\y  - \frac{1}{2} \log | K_\phi + \diag(\bs_\phi^2) |  \\ 
  &\hspace{-10mm} \qquad + \log \mathrm{Gamma}(\sigma_E | \alpha_E, \beta_E) + \log \mathrm{Gamma}(\sigma_S | \alpha_S,\beta_S), 
  \end{split}\label{eq:mll}
\end{align}
which automatically balances model fit (the square term) and the model complexity (the determinant) to avoid overfitting \citep{rasmussen}. The parameters can be optimised by maximising the marginal log likelihood~\eqref{eq:mll} using gradient ascent, since the marginal likelihood can be differentiated analytically (see Supplementary Equations~S1 and S2). We utilised a limited-memory projected quasi-Newton algorithm (\texttt{minConf\_TMP}\footnote{\texttt{http://www.cs.ubc.ca/\textasciitilde schmidtm/Software/minConf.html}}), described by \citep{schmidt2009optimizing}. 

\subsection{Evaluation criteria} \label{sec:eval}

We chose to evaluate the accuracy of our predictions using the same metrics that have been used by many others -- correlation $\rho$ between the predicted and experimentally measured $\DDG$ values \citep{capriotti,pop2,kellogg,mcsm,potapov} and the root mean square error ($\rmse$) \citep{pop2,duet,mcsm}, which are determined in the Supplementary Equations S3 and S4. We use marginal likelihood maximisation to infer model parameters and perform cross-validation to evaluate the model performance on test data. Below we only report evaluation metrics obtained from the test sets not used at any stage of the model learning or data transformation sampling.

\section{Results}

In this section we evaluate the performance of mGPfusion on predicting stability effects of mutations, and compare it to the state-of-the-art prediction methods mCSM, PoPMuSiC and Rosetta. Rosetta is a molecular modelling software whose \texttt{ddg\_monomer} module can directly simulate the stability changes $\DDG$ of a protein upon mutations. PoPMuSic and mCSM are machine learning models that predict stability based on protein variant features. We run Rosetta locally, and use mCSM and PoPMuSiC models through their web servers\footnote{\texttt{biosig.unimelb.edu.au/mcsm} and \\ \texttt{omictools.com/popmusic-tool}}. This may give these methods an advantage over mGPfusion since parts of our testing data were likely used within their training data. 

We compare four different variants of our method: mGPfusion that uses both simulated data and MKL, ``mGPfusion, only B62'' that uses simulated data but incorporates only one kernel matrix (BLOSUM62 substitution matrix), mGP model that uses MKL but does not use simulated data, and ``mGP, only B62'' that uses only the base GP model but does not incorporate simulated data and uses only the BLOSUM62 substitution matrix. 
In addition, we experiment on transforming Rosetta predictions with the Bayesian scaling. We perform the experiments for the 15 proteins separately using either position or mutation level (leave-one-out) cross-validation regarding the methods mGP, mGPfusion and the Bayesian scaling of Rosetta.
\citet{mcsm} used protein and position level cross-validation to evaluate their model. In protein level cross-validation all mutations in a protein are either in the test or training set exclusively. When we train our model using protein level cross-validation, we use no experimental data and rely only on the simulated data. Position level cross-validation is defined so that all mutations in a position are either in the test or training set exclusively. However, datasets in \citet{mcsm} contained only point mutations and therefore we had to extend the definition to also include multiple mutations. In position level cross-validation we train one model for each position using only the part of data that has a wild-type residue in that position. Therefore, in position level cross-validation we construct a test set that contains all protein variants that have a mutation at position $p$ and use as training set all the protein variants that have a wild-type residue at that position.
\citet{pop2} evaluated their models by randomly selecting training and test sets so that each mutation was exclusively in one of the sets, but both sets could contain mutations from the same position of the same protein. We call this mutation level cross-validation. When we use all available experimental data with mutation level cross-validation, this corresponds to leave-one-out cross-validation.

\subsection{Predicting point mutations} \label{sec:comparison}

{\begin{table*}[tb]
\center
{\resizebox{\linewidth}{!}{
\begin{tabular}{@{} l | rcl | rcl | rcl || rcl | rcl | rcl @{}}
\toprule 
  & \multicolumn{9}{c||}{Correlation $\rho$} & \multicolumn{9}{c}{$\rmse$}\\
  & \multicolumn{3}{c|}{Point mutations} & \multicolumn{3}{c|}{Multiple mutations} & \multicolumn{3}{c||}{All mutations} & \multicolumn{3}{c|}{Point mutations} & \multicolumn{3}{c|}{Multiple mutations} & \multicolumn{3}{c}{All mutations} \\
\cline{2-19}
& \multicolumn{3}{c|}{ cross-validation level }& \multicolumn{3}{c|}{ cross-validation level }& \multicolumn{3}{c||}{ cross-validation level }& \multicolumn{3}{c|}{ cross-validation level }& \multicolumn{3}{c|}{ cross-validation level }& \multicolumn{3}{c}{ cross-validation level }\\
Method & \:\,  mut. & pos. & \multicolumn{1}{l|}{ prot.} & \:\,  mut. & pos. & \multicolumn{1}{l|}{ prot.} & \:\,  mut. & pos. & prot. & \:\,  mut. & pos. & \multicolumn{1}{l|}{ prot.} & \:\,  mut. &  pos. & \multicolumn{1}{l|}{ prot.} & \:\,  mut. & pos. & prot.\\
\hline
mGPfusion & \bf{0.81} & \bf{0.70} & 0.56 & \bf{0.88} & 0.61 & 0.49 & \bf{0.83} & 0.64 & 0.52 & 1.07 & \bf{1.26} & 1.61 & \bf{1.33} & 2.45 & 2.53 & \bf{1.13} & 1.87 & 1.84   \\
mGPfusion, only B62 & 0.79 & 0.69 & 0.56 & 0.86 & \bf{0.64} & 0.50 & 0.82 & \bf{0.66} & 0.52 & 1.11 & 1.30 & 1.62  & 1.43 & \bf{2.40} & 2.50 & 1.18 & \bf{1.85} & 1.84 \\
mGP & 0.81 & 0.51 & \:\:\:\:-  & 0.86 & 0.52 & \:\:\:\:- & 0.83 & 0.50 & \:\:\:\:- & \bf{1.04} & 1.54 & \:\:\:\:- & 1.44 & 2.65 & \:\:\:\:- & 1.14 & 2.09 & \:\:\:\:- \\
mGP, only B62 & 0.76 & 0.34 & \:\:\:\:- & 0.86 & 0.55 & \:\:\:\:- & 0.80 & 0.49 & \:\:\:\:- & 1.26 & 1.95 & \:\:\:\:- & 1.45 & 2.56 & \:\:\:\:- & 1.30 & 2.23 &  \:\:\:\:- \\
Rosetta scaled & 0.65 & 0.63 & \:\:\:\:- & 0.51 & 0.39 & \:\:\:\:- & 0.60 & 0.48 & \:\:\:\:-  & 1.35 & 1.38 & \:\:\:\:-  & 2.49 & 2.99 & \:\:\:\:-  & 1.66 & 2.22 & \:\:\:\:- \\\midrule
\multicolumn{19}{l}{Predictions from off-the-shelf implementations with no cross-validation}\\\midrule
Rosetta & \multicolumn{3}{c|}{0.55} &\multicolumn{3}{c|}{0.40} & \multicolumn{3}{c||}{0.49} & \multicolumn{3}{c|}{1.63} & \multicolumn{3}{c|}{2.74} & \multicolumn{3}{c}{1.92} \\
mCSM     & \multicolumn{3}{c|}{0.61} & \multicolumn{3}{c|}{-}  & \multicolumn{3}{c||}{-} & \multicolumn{3}{c|}{1.40} & \multicolumn{3}{c|}{-}  & \multicolumn{3}{c}{-} \\
PoPMuSiC & \multicolumn{3}{c|}{0.64} & \multicolumn{3}{c|}{-}  & \multicolumn{3}{c||}{-} & \multicolumn{3}{c|}{1.37} & \multicolumn{3}{c|}{-}  & \multicolumn{3}{c}{-} \\\bottomrule
\end{tabular}}}
\captionsetup{width=\linewidth}
\caption{Comparison of different methods on the $15$ protein dataset with respect to $\rho$ and $\rmse$. Mutation, position, and protein are referred to as mut., pos., and prot., respectively. Predictions from off-the-shelf implementations of Rosetta, mCSM and PoPMuSiC are used directly without cross-validation.} \label{tab:comparison}
\end{table*}}

Table~\ref{tab:comparison} summarises the average prediction performance over all $15$ proteins for all compared methods, types of mutations and cross-validation types. We first compare the performances on single point mutations, where mGPfusion and mGP achieve the highest performance with $\rho = 0.81$ and $\rmse = 1.07$ kcal/mol, and $\rho = 0.81$ and $\rmse = 1.04$ kcal/mol, respectively with mutation level cross-validation. With only one kernel utilising the BLOSUM62  matrix instead of MKL, the performance decreases slightly, but the competing methods are still outperformed, as mCSM achieves $\rho = 0.64$ and $\rmse = 1.37$ kcal/mol, PoPMuSic $\rho = 0.61$ and Rosetta $\rho = 0.55$. Applying Bayesian scaling on Rosetta simulator improves the performance of standard Rosetta from $\rho = 0.55$ to $\rho = 0.65$ and decreases the $\rmse$ from $1.63$ kcal/mol to $1.35$ kcal/mol, which is interestingly even slightly better than the performances of mCSCM and PoPMuSiC. 

With position level cross-validation mGPfusion achieves the highest performance of $\rho = 0.70$ and $\rmse = 1.26$ kcal/mol, likely due to having still access to simulated variants from that position, since they are always available to the learner. Without simulation data, the baseline machine learning model mGP performance decreases to $\rho = 0.51$ and $\rmse = 1.54$ kcal/mol, thus demonstrating the importance of the data fusion. Cross-validation could not be performed for the off-the-shelf methods mCSM and PoPMuSiC. Even still, mGPfusion (trained with one or multiple kernels) outperforms competing state-of-the-art methods and achieves markedly higher prediction performance as quantified by both mutation and position level cross-validations. Also mGP outperforms these methods when quantified by mutation level cross-validation.
With protein level cross-validation mGPfusion achieves slightly better results than Rosetta.

\subsection{Predicting multiple mutations}

\begin{figure}[H]
\includegraphics[width=\linewidth]{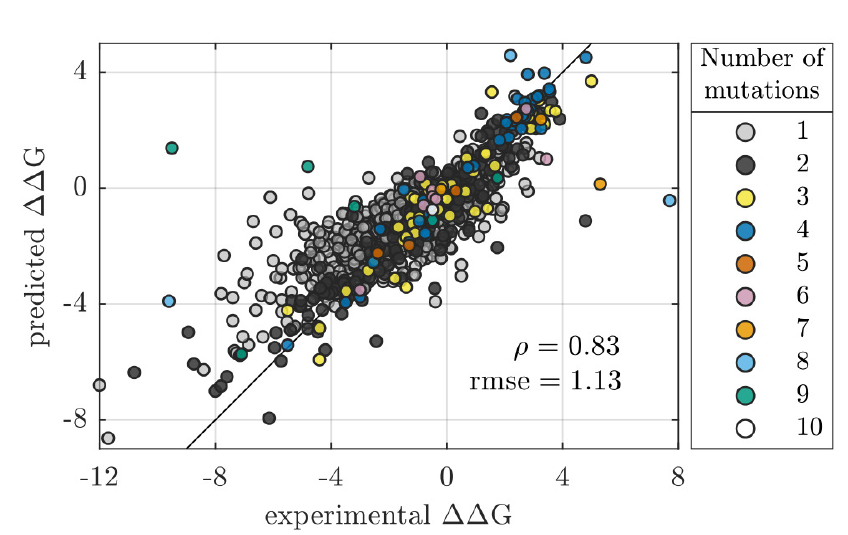}
\caption{Scatter plot for the mutation level (leave-one-out) predictions made for all 15 proteins (See Table~\ref{tab:thermal}). The colour indicates the number of simultaneous mutations.}
\label{fig:scatter}
\end{figure}

Next, we tested stability prediction accuracies for variants containing either single or multiple mutations. Figure~\ref{fig:scatter} shows a scatter plot of mGPfusion predictions for all $1537$ single and multiple mutation variants (covering all $15$ proteins) against the experimental $\DDG$ values using the mutation level (leave-one-out) cross-validation. The points are coloured by the number of simultaneous mutations in the variants, with $326$ variants having at least $2$ mutations (See Table~\ref{tab:thermal}).  Figure~\ref{fig:scatter} illustrates the mGPfusion's overall high accuracy of $\rho = 0.83$ and $\rmse = 1.13$ kcal/mol on both single and multiple mutations (See Table~\ref{tab:comparison}). Scatter plots for the individual proteins can be found in Supplementary Figure~S3. \citet{pop2} suggested that considering the predictive power after removal of most badly predicted stability effects of mutations may give more relevant evaluation, as some of the experimental measurements may have been made in non-physiological conditions or affected by significant error, associated with a poorly resolved structure, or indexed incorrectly in the database. They thus reported correlation and rmse of the predictions after excluding 10 \% of the predictions with most negative impacts on the correlation coefficient. \citet{mcsm} also reported their accuracy after 10 \% outlier removal. If we remove the 10\% worst predicted stability effects from the combined predictions, we achieve correlation $\rho$ of $0.92$ and $\rmse$ of $0.67$ kcal/mol. We report these results for all the methods in Supplementary Table~S3 and also present the error distribution in Supplementary Figure~S5.

The high accuracy is retained for variants with multiple mutations as well ($\rho = 0.88$ and $\rmse = 1.33$ kcal/mol, see Tables~\ref{tab:comparison} and S2). Table~\ref{tab:multimutations} lists mGPfusion's $\rmse$ for different number of simultaneous mutations. The model accuracy in fact improves up to $6$ mutations. This is explained by the training set often containing the same single point mutations that appear in variants with multiple mutations. The model can then infer the combined effect of pointwise mutations. The model seems to fail when predicting the effects of 7-9 simultaneous mutations. Most of these mutations (8/12) are for Ribonuclease (1RGG) and their effects seem to be exceptionally difficult to predict. This may be because only few of the point mutations that are part of the multiple mutations are present in the training data. However, these mutations seem to be exceptionally difficult to predict for Rosetta as well, which could indicate that the experimental measurements concerning these mutations are not quite accurate. 
PoPMuSiC and mCSM are unable to predict multiple mutations, while Rosetta supports them, but its $\rmse$ accuracy decreases already with two mutations.

\begin{table}[H]
\resizebox{\linewidth}{!}{
\begin{tabular}{@{}lcccccccccc@{}}\toprule 
          mutations &    1 &    2 &    3 &    4 &    5 &    6 &    7 &    8 &    9 &   10 \\
         occurences & 1211 &  207 &   52 &   42 &    4 &    8 &    3 &    3 &    6 &    1 \\\midrule
          mGPfusion & 1.07 & 1.06 & 0.80 & 0.51 & 0.40 & 1.01 & 3.02 & 5.89 & 5.16 & 0.25 \\
          mGPfusion, only B62\!\! & 1.11 & 1.12 & 0.77 & 0.59 & 0.29 & 1.14 & 3.00 & 6.78 & 5.56 & 0.11 \\
                mGP & 1.04 & 1.03 & 0.61 & 0.50 & 0.18 & 0.92 & 3.23 & 6.18 & 6.75 & 0.08 \\
      mGP, only B62 & 1.26 & 0.96 & 0.65 & 0.83 & 0.26 & 1.14 & 2.95 & 6.90 & 6.57 & 0.05 \\
     Rosetta scaled & 1.35 & 2.10 & 1.92 & 2.94 & 2.29 & 2.32 & 2.93 & 6.75 & 7.28 & 2.69 \\
            Rosetta & 1.63 & 2.27 & 2.11 & 3.78 & 2.93 & 2.21 & 2.92 & 5.80 & 7.45 & 3.42 \\\bottomrule
\end{tabular}}{}
\caption{Root-mean-square errors for different number of simultaneous mutations for all 15 proteins, with models trained by leave-one-out cross-validation. Rosetta is added for comparison.} \label{tab:multimutations}
\end{table}

With multiple mutations, the decrease in performance between the position and mutation level cross-validations becomes clearer than with single mutations. With the position level cross-validation the stability effects of multiple mutations are predicted multiple times, which partly explains this loss of accuracy. For example, the effects of mutants with nine different simultaneous mutations, which were the most difficult cases in the mutation level cross-validation, are predicted nine times. Surprisingly, mGPfusion trained with protein level cross-validation achieves higher correlation and smaller errors than Rosetta; mGPfusion utilising simulated $\DDG$ values for only single mutations, can predict the stability effects of multiple mutations better than Rosetta.

\subsection{Uncertainty of the predictions}\label{sec:uncertainty}

Gaussian processes provide a mean $\mu(\x)$ and a standard deviation $\sigma(\x)$ for the stability prediction of a protein variant $\x$. The standard deviation allows estimation of the prediction accuracy even without test data. Figure~\ref{fig:pipeline}~h) visualises the uncertainty of a few predictions made for the protein G (1PGA) when mutation level cross-validation is used. The estimated standard deviation allows a user to automatically identify low quality predictions that can appear e.g.\ in parts of the input protein space from which less data is included in model training. Conversely, in order to minimise the amount of uncertainty in the mGPfusion predictions, estimated standard deviation can be used to guide next experiments. The probabilistic nature of the predictions also admits an alternative error measure of negative log probability density (NLPD) $\mathrm{nlpd} = - \sum_{i=1}^N \log p(y_i | \mu(\x_i), \sigma^2(\x_i))$, which can naturally take into account the prediction variance.

\subsection{Effect of training set size}

The results presented in Sections~\ref{sec:comparison}--\ref{sec:uncertainty} used all available data for training with cross-validation to obtain unbiased performance measures. The inclusion of thousands of simulated variants allows the model to learn accurate models with less experimentally measured variants. Hence, we study how the mGPfusion model with or without simulated data performs with reduced number of experimental observations. To facilitate this, we randomly selected subsets of experimental data of size $0$, $10$, $20$, and so on. We learned the mGP and mGPfusion models with these reduced experimental data sets while always using the full simulated data sets. This also allows us to estimate how the models work with different number of cross-validation folds. For example, the point of a learning curve which utilises $2/3$ or $4/5$ of the training data correspond to an average of multiple 3-fold or 5-fold cross-validations, respectively.

\begin{figure}[H]
\centerline{\includegraphics[width=\linewidth]{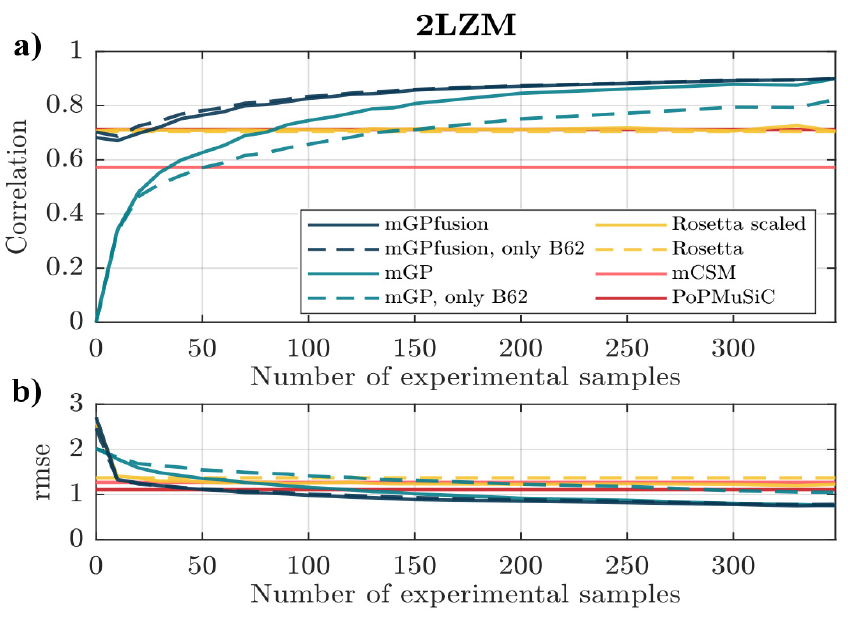}}
\caption{a) Correlation and b) root mean square error of predictions made by models with different number of experimental training samples for T4 Lysozyme (2LZM). The results of Rosetta, mCSM and PoPMuSiC are invariant to training data (because mCSM and PoPMuSiC are pre-trained), and are thus constant lines. For both figures, an average of 100 randomly selected training sets is taken at each point. }
\label{fig:curve_2LZM}
\end{figure}

The learning curve in Figure~\ref{fig:curve_2LZM}a) shows how the averaged correlation for protein 2LZM improves when the size of the experimental data set increases. The right-most values at $N=348$ are obtained with leave-one-out cross-validation. The inclusion of simulated data in mGPfusion (dark blue line) consistently improves the performance of mGP, which is trained without simulated data. Figure~\ref{fig:curve_2LZM}b) illustrate the difference in root mean square error. Learning curves for all proteins listed in Table~\ref{tab:thermal} can be found from the Supplementary Figures~S6-S8.
When the number of experimental samples is zero, the mGPfusion model is trained solely using the simulated data with scaling $0.57 y^S$, and the mGP model predicts the stability effect of every mutation as zero. The last point on the learning curves is obtained with mutation level cross-validation (see Tables~\ref{tab:comparison} and S2).

\subsection{Effect of data fusion and multiple substitution matrices}

In the beginning of the learning curves, when only little training data is available, mGPfusion quite consistently outperforms the mGP model, demonstrating that the additional simulated data improves the prediction accuracy. However, when more training data becomes available, the performance of mGP model is almost as good or sometimes even better than the performance of the mGPfusion model. This shows that if enough training data is available, it is not necessary to simulate additional data in order to obtain accurate predictions. Table~\ref{tab:comparison} also shows, that the data fusion can compensate the lack of relevant training data -- with the mGPfusion models that utilise the additional data, the decrease in accuracy is smaller when position level cross-validation is used instead of mutation level cross-validation, than with the mGP models.

The varying weights for the base kernels between different proteins (shown in Figure~\ref{fig:kernelweights}) already illustrated that different proteins benefit from different similarity measures for amino acid substitutions. The learning curves also support this observation -- with some of the proteins mGPfusion model trained with only one kernel that utilises BLOSUM62, provides approximately as good results as the mGPfusion model trained with multiple kernels. However, with many of the proteins, utilising just BLOSUM62 does not seem to be sufficient and the accuracy of the model can be improved by using different substitution matrices. Prior knowledge of appropriate substitution models for each protein could enable creation of accurate prediction models with just one substitution model, but the MKL seems to be a good tool for selecting suitable substitution models when such knowledge is not available. 
It seems that the data fusion and number or relevance of used substitution matrices can compensate each other -- the learning curves show, that the difference between mGPfusion models trained with one or multiple kernels is smaller than the difference between the mGP models utilising one or multiple kernels. This indicates that if additional simulated data is exploited, the use of multiple or appropriate substitution models is not as important than without the data fusion. On the other hand, if data fusion is not applied, the use of MKL can more significantly improve the accuracy of the mGP model.

\subsection{Effect of the Bayesian transformation on Rosetta}

The Bayesian scaling of simulated Rosetta values, proposed in Section~\ref{sec:scaling}, improves the match of Rosetta simulated values to empirical $\DDG$ values even without using the Gaussian process framework. The Bayesian scaling improves the performance of standard Rosetta simulations from $\rho = 0.55$ and $\rmse = 1.63$ kcal/mol to $\rho = 0.65$ and $\rmse = 1.35$ kcal/mol (see Table \ref{tab:comparison} and Supplementary Table S2). This shows that the scaling proposed by \citet{kellogg} indeed is not always the optimal scaling and significant improvements can be gained by optimising the scaling using a set of training data.

Figure~\ref{fig:pipeline}~g) visualises the Bayesian scaling for protein 1PGA, where the very destabilising $\DDG$ values are dampened by the scaling (black dots) to less extreme values by matching the scaled simulated values to the experimental points (blue circles). The black dots along the scaling curve indicate the grid of point mutations after transformation. The scaling variance $\bsRT^2$ is indicated by the green region's vertical width, and on the right panel. The scaling tends to dampen very small values into less extreme stabilities, while it also estimates higher uncertainties for stability values further away from $\DDG = 0$. However, the scalings vary between different proteins, as can be seen from the transformations for each of the 15 proteins presented in Supplementary Figure~S9.

\section{Conclusions}

We present a novel method mGPfusion for predicting stability effects of both single and multiple simultaneous mutations. mGPfusion utilises structural information in form of contact maps and integrates that with amino acid residues and combines both experimental and comprehensive simulated measurements of mutations' stability effects. In contrast to earlier general-purpose stability models, mGPfusion model is protein-specific by design, which improves the accuracy but necessitates having a set of experimental measurements from the protein. In practise small datasets of 10--20 experimental observations were found to provide state-of-the-art accuracy models when supplemented by large simulation datasets. 

An important advantage over most state-of-the-art machine learning methods is that mGPfusion is able to predict the effects of multiple simultaneous mutations in addition to single point mutations. Our experiments show that mGPfusion is reliable in predicting up to six simultaneous mutations in our dataset. Furthermore, the Gaussian process framework provide a way to estimate the (un)certainty of the predictions even without a separate test set. We additionally proposed a novel Bayesian scaling method to re-calibrate simulated protein stability values against experimental observations. This is a crucial part of the mGPfusion model, and also alone improved protein-specific Rosetta stability predictions by calibrating them using experimental data.

mGPfusion is best suited for a situation, where a protein is thoroughly experimented on and accurate predictions for stability effects upon mutations are needed. It takes some time to set up the framework and train the model, but after that new predictions can be made in fractions of a second. The most time-consuming part is running the simulations with Rosetta, at least when the most accurate protocol 16 is used. Simulating all 19 possible point mutations for one position took about 12 hours, but simulations for different positions can be run on parallel. The time needed for training the prediction model depends on the amount of experimental and simulated training data. With no simulated data, the training time ranged from few seconds to few minutes. With data fusion and a single kernel, the training time was under an hour. With data fusion and MKL with 21 kernels, the training time was from a few minutes to a day.

\section*{Acknowledgements}
We acknowledge the computational resources provided by the Aalto Science-IT.

\section*{Funding}
This work has been supported by the Academy of Finland Center of Excellence in Systems Immunology and Physiology, the Academy of Finland grants no. 260403 and 299915, and the Finnish Funding Agency for Innovation Tekes (grant no 40128/14, Living Factories). \vspace*{-12pt}

\bibliographystyle{natbib}

\begin{thebibliography}{}

\bibitem[Alberts {\em et~al.}(2007)Alberts, Johnson, Lewis, Raff, Roberts, and
  Walter]{alberts}
Alberts, B., Johnson, A., Lewis, J., Raff, M., Roberts, K., and Walter, P.
  (2007).
\newblock {\em Molecular biology of the cell\/}.
\newblock Garland Science, 5 edition.

\bibitem[Anslyn and Dougherty(2006)Anslyn and Dougherty]{mpoc}
Anslyn, E.~V. and Dougherty, D.~A. (2006).
\newblock {\em Modern physical organic chemistry\/}.
\newblock University Science Books.

\bibitem[Berman {\em et~al.}(2000)Berman, Westbrook, Feng, Gilliland, Bhat,
  Weissig, Shindyalov, and Bourne]{pdb2000}
Berman, H.~M., Westbrook, J., Feng, Z., Gilliland, G., Bhat, T., Weissig, H.,
  Shindyalov, I.~N., and Bourne, P.~E. (2000).
\newblock The protein data bank.
\newblock {\em Nucleic acids research\/}, {\bf 28}(1), 235--242.

\bibitem[Bommarius {\em et~al.}(2011)Bommarius, Blum, and Abrahamson]{industry}
Bommarius, A.~S., Blum, J.~K., and Abrahamson, M.~J. (2011).
\newblock Status of protein engineering for biocatalysts: how to design an
  industrially useful biocatalyst.
\newblock {\em Current opinion in chemical biology\/}, {\bf 15}(2), 194--200.

\bibitem[Branden and Tooze(1999)Branden and Tooze]{proteinStructure}
Branden, C. and Tooze, J. (1999).
\newblock {\em Introduction to protein structure\/}.
\newblock Garland, 2 edition.

\bibitem[Capriotti {\em et~al.}(2005a)Capriotti, Fariselli, and
  Casadio]{capriotti}
Capriotti, E., Fariselli, P., and Casadio, R. (2005a).
\newblock {I-Mutant2.0}: predicting stability changes upon mutation from the
  protein sequence or structure.
\newblock {\em Nucleic acids research\/}, {\bf 33}(suppl 2), W306--W310.

\bibitem[Capriotti {\em et~al.}(2005b)Capriotti, Fariselli, Calabrese, and
  Casadio]{imutant2}
Capriotti, E., Fariselli, P., Calabrese, R., and Casadio, R. (2005b).
\newblock Predicting protein stability changes from sequences using support
  vector machines.
\newblock {\em Bioinformatics\/}, {\bf 21}(suppl 2), ii54--ii58.

\bibitem[Capriotti {\em et~al.}(2008)Capriotti, Fariselli, Rossi, and
  Casadio]{threestate}
Capriotti, E., Fariselli, P., Rossi, I., and Casadio, R. (2008).
\newblock A three-state prediction of single point mutations on protein
  stability changes.
\newblock {\em BMC bioinformatics\/}, {\bf 9}(2).

\bibitem[Chen {\em et~al.}(2013)Chen, Lin, and Chu]{istable}
Chen, C.-W., Lin, J., and Chu, Y.-W. (2013).
\newblock {iStabl}e: off-the-shelf predictor integration for predicting protein
  stability changes.
\newblock {\em BMC bioinformatics\/}, {\bf 14}(2).

\bibitem[Cheng {\em et~al.}(2006)Cheng, Randall, and Baldi]{mupro}
Cheng, J., Randall, A., and Baldi, P. (2006).
\newblock Prediction of protein stability changes for single-site mutations
  using support vector machines.
\newblock {\em Proteins: Structure, Function, and Bioinformatics\/}, {\bf
  62}(4), 1125--1132.

\bibitem[Cherry and Fidantsef(2003)Cherry and Fidantsef]{directedenzymes}
Cherry, J.~R. and Fidantsef, A.~L. (2003).
\newblock Directed evolution of industrial enzymes: an update.
\newblock {\em Current opinion in biotechnology\/}, {\bf 14}(4), 438--443.

\bibitem[Cichonska {\em et~al.}(2017)Cichonska, Ravikumar, Parri, Timonen,
  Pahikkala, Airola, Wennerberg, Rousu, and Aittokallio]{cichonska17}
Cichonska, A., Ravikumar, B., Parri, E., Timonen, S., Pahikkala, T., Airola,
  A., Wennerberg, K., Rousu, J., and Aittokallio, T. (2017).
\newblock Computational-experimental approach to drug-target interaction
  mapping: A case study on kinase inhibitors.
\newblock {\em PLoS computational biology\/}, {\bf 13}(8), e1005678.

\bibitem[Dehouck {\em et~al.}(2009)Dehouck, Grosfils, Folch, Gilis, Bogaerts,
  and Rooman]{pop2}
Dehouck, Y., Grosfils, A., Folch, B., Gilis, D., Bogaerts, P., and Rooman, M.
  (2009).
\newblock Fast and accurate predictions of protein stability changes upon
  mutations using statistical potentials and neural networks: {PoPMuSiC-2.0}.
\newblock {\em Bioinformatics\/}, {\bf 25}(19), 2537--2543.

\bibitem[Folkman {\em et~al.}(2014)Folkman, Stantic, and Sattar]{folkman}
Folkman, L., Stantic, B., and Sattar, A. (2014).
\newblock Feature-based multiple models improve classification of
  mutation-induced stability changes.
\newblock {\em BMC genomics\/}, {\bf 15}(Suppl 4).

\bibitem[Giguere {\em et~al.}(2013)Giguere, Marchand, Laviolette, Drouin, and
  Corbeil]{giguere13}
Giguere, S., Marchand, M., Laviolette, F., Drouin, A., and Corbeil, J. (2013).
\newblock Learning a peptide-protein binding affinity predictor with kernel
  ridge regression.
\newblock {\em BMC bioinformatics\/}, {\bf 14}(1), 82.

\bibitem[Giollo {\em et~al.}(2014)Giollo, Martin, Walsh, Ferrari, and
  Tosatto]{neemo}
Giollo, M., Martin, A.~J., Walsh, I., Ferrari, C., and Tosatto, S.~C. (2014).
\newblock {NeEMO}: a method using residue interaction networks to improve
  prediction of protein stability upon mutation.
\newblock {\em BMC genomics\/}, {\bf 15}(4), 1.

\bibitem[Henikoff and Henikoff(1992)Henikoff and Henikoff]{henikoff1992}
Henikoff, S. and Henikoff, J.~G. (1992).
\newblock Amino acid substitution matrices from protein blocks.
\newblock {\em Proceedings of the National Academy of Sciences\/}, {\bf
  89}(22), 10915--10919.

\bibitem[Kawashima {\em et~al.}(2008)Kawashima, Pokarowski, Pokarowska,
  Kolinski, Katayama, and Kanehisa]{aaindex}
Kawashima, S., Pokarowski, P., Pokarowska, M., Kolinski, A., Katayama, T., and
  Kanehisa, M. (2008).
\newblock {AAindex}: amino acid index database, progress report 2008.
\newblock {\em Nucleic acids research\/}, {\bf 36}(suppl 1), D202--D205.

\bibitem[Kellogg {\em et~al.}(2011)Kellogg, Leaver-Fay, and Baker]{kellogg}
Kellogg, E.~H., Leaver-Fay, A., and Baker, D. (2011).
\newblock Role of conformational sampling in computing mutation-induced changes
  in protein structure and stability.
\newblock {\em Proteins: Structure, Function, and Bioinformatics\/}, {\bf
  79}(3), 830--838.

\bibitem[Kirk {\em et~al.}(2002)Kirk, Borchert, and
  Fuglsang]{industrialenzymes}
Kirk, O., Borchert, T.~V., and Fuglsang, C.~C. (2002).
\newblock Industrial enzyme applications.
\newblock {\em Current opinion in biotechnology\/}, {\bf 13}(4), 345--351.

\bibitem[Kumar {\em et~al.}(2006)Kumar, Bava, Gromiha, Prabakaran, Kitajima,
  Uedaira, and Sarai]{protherm}
Kumar, M.~S., Bava, K.~A., Gromiha, M.~M., Prabakaran, P., Kitajima, K.,
  Uedaira, H., and Sarai, A. (2006).
\newblock {ProTherm} and {ProNIT}: thermodynamic databases for proteins and
  protein--nucleic acid interactions.
\newblock {\em Nucleic Acids Research\/}, {\bf 34}(suppl 1), D204--D206.

\bibitem[Leaver-Fay {\em et~al.}(2011)Leaver-Fay, Tyka, Lewis, Lange, Thompson,
  Jacak, Kaufman, Renfrew, Smith, Sheffler, {\em et~al.}]{rosetta3}
Leaver-Fay, A., Tyka, M., Lewis, S.~M., Lange, O.~F., Thompson, J., Jacak, R.,
  Kaufman, K., Renfrew, P.~D., Smith, C.~A., Sheffler, W., {\em et~al.} (2011).
\newblock {ROSETTA3}: an object-oriented software suite for the simulation and
  design of macromolecules.
\newblock {\em Methods in enzymology\/}, {\bf 487}, 545.

\bibitem[Liu and Kang(2012)Liu and Kang]{gradingaaprop15}
Liu, J. and Kang, X. (2012).
\newblock Grading amino acid properties increased accuracies of single point
  mutation on protein stability prediction.
\newblock {\em BMC bioinformatics\/}, {\bf 13}(1), 1.

\bibitem[Menchetti {\em et~al.}(2005)Menchetti, Costa, and Frasconi]{wdk}
Menchetti, S., Costa, F., and Frasconi, P. (2005).
\newblock Weighted decomposition kernels.
\newblock In {\em Proceedings of the 22nd international conference on Machine
  learning\/}, pages 585--592. ACM.

\bibitem[Pace and Scholtz(1997)Pace and Scholtz]{measuring}
Pace, C.~N. and Scholtz, J.~M. (1997).
\newblock Measuring the conformational stability of a protein.
\newblock {\em Protein structure: A practical approach\/}, {\bf 2}, 299--321.

\bibitem[Pace and Shaw(2000)Pace and Shaw]{lem}
Pace, C.~N. and Shaw, K.~L. (2000).
\newblock Linear extrapolation method of analyzing solvent denaturation curves.
\newblock {\em Proteins: Structure, Function, and Bioinformatics\/}, {\bf
  41}(S4), 1--7.

\bibitem[Pires {\em et~al.}(2014a)Pires, Ascher, and Blundell]{duet}
Pires, D.~E., Ascher, D.~B., and Blundell, T.~L. (2014a).
\newblock {DUET}: a server for predicting effects of mutations on protein
  stability using an integrated computational approach.
\newblock {\em Nucleic acids research\/}, page gku411.

\bibitem[Pires {\em et~al.}(2014b)Pires, Ascher, and Blundell]{mcsm}
Pires, D.~E., Ascher, D.~B., and Blundell, T.~L. (2014b).
\newblock {mCSM}: predicting the effects of mutations in proteins using
  graph-based signatures.
\newblock {\em Bioinformatics\/}, {\bf 30}(3), 335--342.

\bibitem[Potapov {\em et~al.}(2009)Potapov, Cohen, and Schreiber]{potapov}
Potapov, V., Cohen, M., and Schreiber, G. (2009).
\newblock Assessing computational methods for predicting protein stability upon
  mutation: good on average but not in the details.
\newblock {\em Protein Engineering Design and Selection\/}, {\bf 22}(9),
  553--560.

\bibitem[Rapley and Walker(2000)Rapley and Walker]{rapley}
Rapley, R. and Walker, J.~M. (2000).
\newblock {\em Molecular Biology and Biotechnology\/}.
\newblock Royal Society of Chemistry, 4 edition.

\bibitem[Rasmussen and Williams(2006)Rasmussen and Williams]{rasmussen}
Rasmussen, C.~E. and Williams, C. K.~I. (2006).
\newblock {\em Gaussian processes for machine learning\/}.
\newblock The MIT Press.

\bibitem[Sanchez and Demain(2010)Sanchez and Demain]{sanchez2010enzymes}
Sanchez, S. and Demain, A.~L. (2010).
\newblock Enzymes and bioconversions of industrial, pharmaceutical, and
  biotechnological significance.
\newblock {\em Organic Process Research \& Development\/}, {\bf 15}(1),
  224--230.

\bibitem[Schmidt {\em et~al.}(2009)Schmidt, Berg, Friedlander, and
  Murphy]{schmidt2009optimizing}
Schmidt, M.~W., Berg, E., Friedlander, M.~P., and Murphy, K.~P. (2009).
\newblock Optimizing costly functions with simple constraints: A limited-memory
  projected quasi-newton algorithm.
\newblock In {\em International Conference on Artificial Intelligence and
  Statistics\/}, page None.

\bibitem[Shawe-Taylor and Cristianini(2004)Shawe-Taylor and
  Cristianini]{shawetaylor04}
Shawe-Taylor, J. and Cristianini, N. (2004).
\newblock {\em Kernel methods for pattern analysis\/}.
\newblock Cambridge university press.

\bibitem[Tian {\em et~al.}(2010)Tian, Wu, Chu, and Fan]{prethermut}
Tian, J., Wu, N., Chu, X., and Fan, Y. (2010).
\newblock Predicting changes in protein thermostability brought about by
  single- or multi-site mutations.
\newblock {\em BMC bioinformatics\/}, {\bf 11}(1), 1.

\bibitem[Tokuriki and Tawfik(2009)Tokuriki and Tawfik]{tokuriki}
Tokuriki, N. and Tawfik, D.~S. (2009).
\newblock Stability effects of mutations and protein evolvability.
\newblock {\em Current opinion in structural biology\/}, {\bf 19}(5), 596--604.

\bibitem[Tomii and Kanehisa(1996)Tomii and Kanehisa]{tomii}
Tomii, K. and Kanehisa, M. (1996).
\newblock Analysis of amino acid indices and mutation matrices for sequence
  comparison and structure prediction of proteins.
\newblock {\em Protein Engineering, Design and Selection\/}, {\bf 9}(1),
  27--36.

\bibitem[Vishwanathan {\em et~al.}(2010)Vishwanathan, Schraudolph, Kondor, and
  Borgwardt]{vishwanathan2010}
Vishwanathan, S. V.~N., Schraudolph, N.~N., Kondor, R., and Borgwardt, K.~M.
  (2010).
\newblock Graph kernels.
\newblock {\em The Journal of Machine Learning Research\/}, {\bf 11},
  1201--1242.

\bibitem[Wainreb {\em et~al.}(2011)Wainreb, Wolf, Ashkenazy, Dehouck, and
  Ben-Tal]{promaya}
Wainreb, G., Wolf, L., Ashkenazy, H., Dehouck, Y., and Ben-Tal, N. (2011).
\newblock Protein stability: a single recorded mutation aids in predicting the
  effects of other mutations in the same amino acid site.
\newblock {\em Bioinformatics\/}, {\bf 27}(23), 3286--3292.

\end{thebibliography}

\end{multicols}

\newpage

\makeatletter
\renewcommand{\thefigure}{S\@arabic\c@figure}
\makeatletter

\makeatletter
\renewcommand{\theequation}{S\@arabic\c@equation}
\makeatletter

\makeatletter
\renewcommand{\thetable}{S\@arabic\c@table}
\makeatletter

\section*{\center \huge Supplementary material}

\vspace{12ex}
\begin{figure}[!htb]
\centerline{\includegraphics[width=\linewidth]{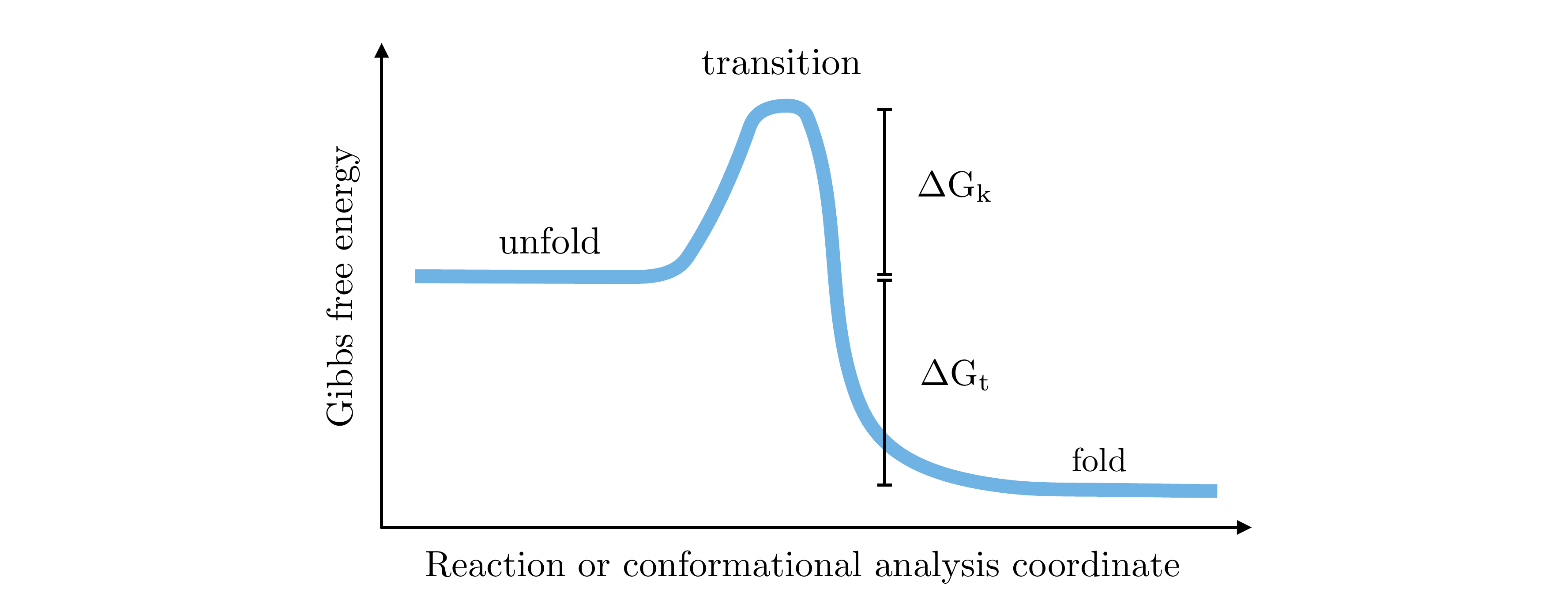}}
\caption{ The stability of a protein is determined by the thermodynamic and kinetic stabilities, $\DG_\mathrm{t}$ and $\DG_\mathrm{k}$, respectively. We only consider the thermodynamic stability.}
\label{fig:gibbs}
\end{figure}

\vspace{6ex}

\begin{figure}[!htb]
\centerline{\includegraphics[width=\linewidth]{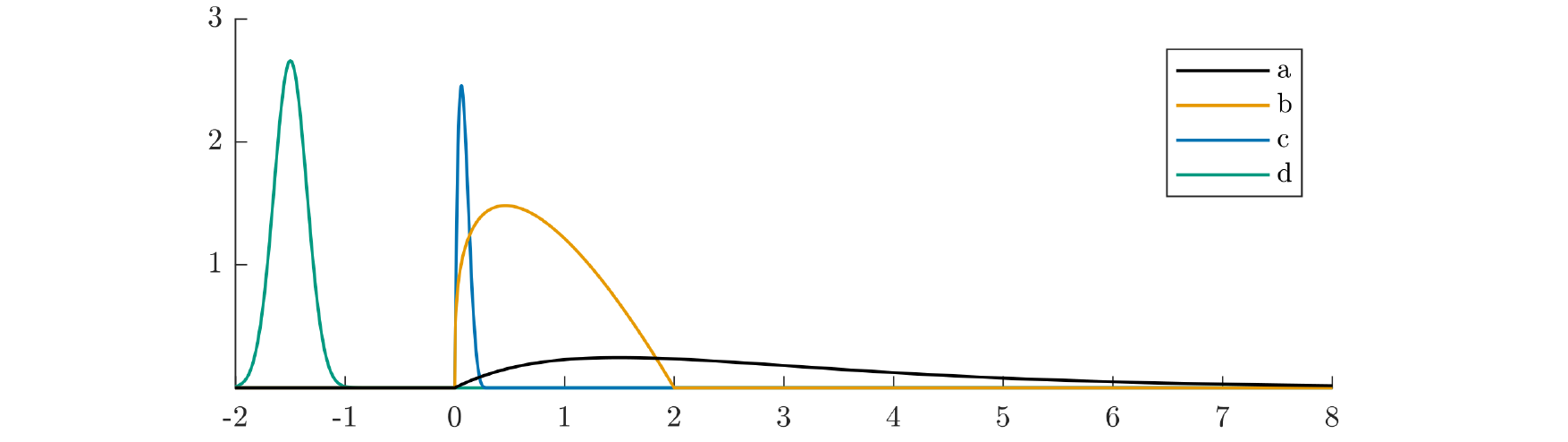}}
\caption{Priors presented by Equation~2. Here $\mu_d=-1.5$, the most likely value for $-a$. Other hyperparameter values are presented in Table~\ref{tab:parameters}.}\label{fig:priors}
\end{figure}

\vspace{6ex}

\begin{table}[h]
\begin{center}
\caption{Values for the hyperparameters used in the priors of $a,b,c,d,\sigma_E$ and $\sigma_S$ presented in Equations~2 and 6, respectively. \vspace{1ex}\label{tab:parameters}} {\begin{tabular}{@{}cccccc@{}}\toprule 
 $a$ & $b$ & $c$ & $d$ & $\sigma_E$ & $\sigma_S$ \\\midrule
 $\alpha_a = 2$  & $\alpha_b = 1.3$ & $\alpha_c = 2$ & $\mu_d = -a$      & $\alpha_E = 2.5$ & $\alpha_S = 50$   \\
 $\beta_a = 1.5$ & $\beta_b = 2$    & $\beta_c = 5$  & $\sigma_d = 0.15$ & $\beta_E = 0.02$ & $\beta_S = 0.007$ \\\bottomrule
\end{tabular}}{}
\end{center}
\end{table}

\newpage

The partial derivatives of the marginal likelihood with respect to the parameters $\boldsymbol{\phi}$ are obtained from Equation~(9) as follows:
\begin{align}
\begin{split}
\frac{\partial}{\partial\phi_j}\log p(\mathbf{y}|X,\boldsymbol{\phi}) =& \frac{1}{2}\mathbf{y}^T K_\phi^{-1}\frac{\partial K_\phi}{\partial\phi_j}K_\phi^{-1}\mathbf{y} - \frac{1}{2}\tr\left(K_\phi^{-1}\frac{\partial K_\phi}{\partial\phi_j}\right)\\
=&\frac{1}{2}\tr\left(\left(\boldsymbol{\alpha\alpha}^T - K_\phi^{-1}\right)\frac{\partial K_\phi}{\partial\theta_j}\right),
\end{split}\label{eq:partialPara}
\end{align}
where $\boldsymbol{\alpha}=K_\phi^{-1}\mathbf{y}$, $K_\phi$ is determined as 
\begin{equation*}
K_\phi = \sum_{m=1}^M w_m K_m^{\gamma_m} +\diag  \begin{pmatrix}
\sigma_0\\
\sigma_E \mathbf{1}_{N_E}\\
\sigma_E \1_{N_E}+\sigma_S \1_{N_S} +t \bs_T \end{pmatrix}^2 
\end{equation*}
and the partial derivatives of $K_\phi$ with respect to the optimised parameters are
\begin{align}
\frac{\partial K_\phi}{\partial\sigma_E} =& \diag \begin{pmatrix}
0\\
2\sigma_E \mathbf{1}_{N_E}\\
2\left(\sigma_E \1_{N_E} +\sigma_S \1_{N_S}+t\bs_T\right) \end{pmatrix}\\
\frac{\partial K_\phi}{\partial\sigma_R} =& \diag \begin{pmatrix}
0\\
\mathbf{0}_{N_E}\\
2\left(\sigma_E \1_{N_E} +\sigma_S \1_{N_S} +t\bs_T\right) \end{pmatrix} \\
\frac{\partial K_\phi}{\partial t} =& \diag \begin{pmatrix}
0\\
\mathbf{0}_{N_E}\\
2\left (\sigma_E \1_{N_E} +\sigma_S \1_{N_S}  +t\right) \bs_T \end{pmatrix}\\
\frac{\partial K_\phi}{\partial w_m} =& K_m^{\gamma_m}\\
\frac{\partial K_\phi}{\partial \gamma_m} =& w_m K_m^{\gamma_m} \log K_m
\label{eq:partials}
\end{align}

Correlation $\rho$ and root-mean-square error $\mathrm{rmse}$ for the predictions are determined as
\begin{align}
\rho &= \frac{\sum_{i=1}^{N_*} (y_i-\bar{\y}) (\mu(\x_i) - \bar{\bmu} ) }{ \sqrt{\sum_{i=1}^{N_*} (y_i - \bar{\y})^2 \sum_{i=1}^{N_*} (\mu(\x_i) - \bar{\bmu} )^2 } } \label{eq:corr} \\
\rmse &= \sqrt{ \frac{1}{N_*} \sum_{i=1}^{N_*} (y_i - \mu(\x_i))^2}, \label{eq:rmse}
\end{align}
where $\bar{\y}$ is the mean of the experimentally measured values, $\mu(\x_i)$ is prediction mean, $\bar{\bmu}$ is the average of all prediction means, and $N_*$ is the number of predictions.

\begin{figure}[!htb]
\centerline{\includegraphics[width=\linewidth]{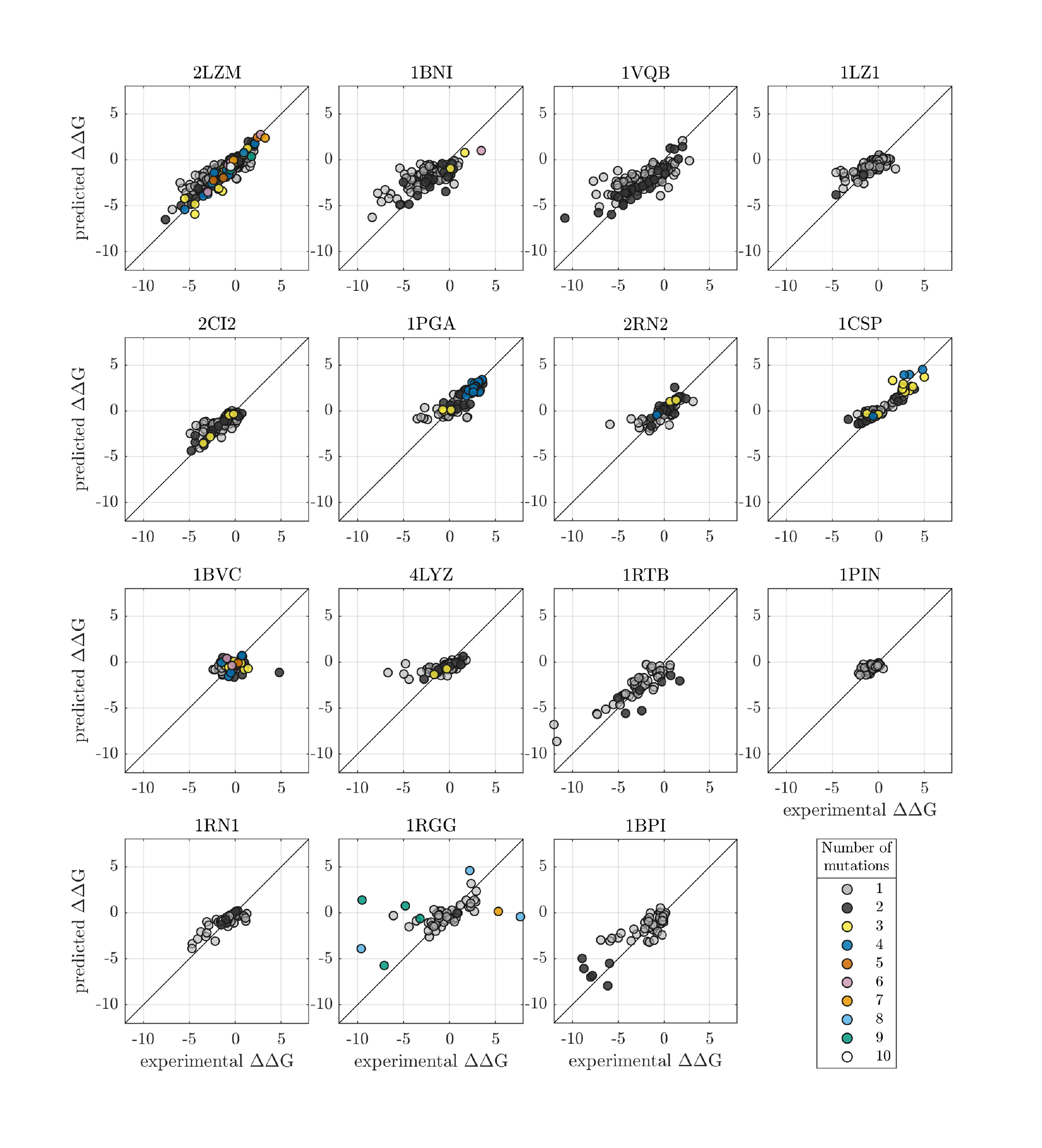}}
\caption{Mutation-level predictions for all 15 proteins presented in Table~1. The predictions are coloured by the number of simultaneous mutations.}\label{fig:loos}
\end{figure}

\begin{figure}[!htb]
\centerline{\includegraphics[width=\linewidth]{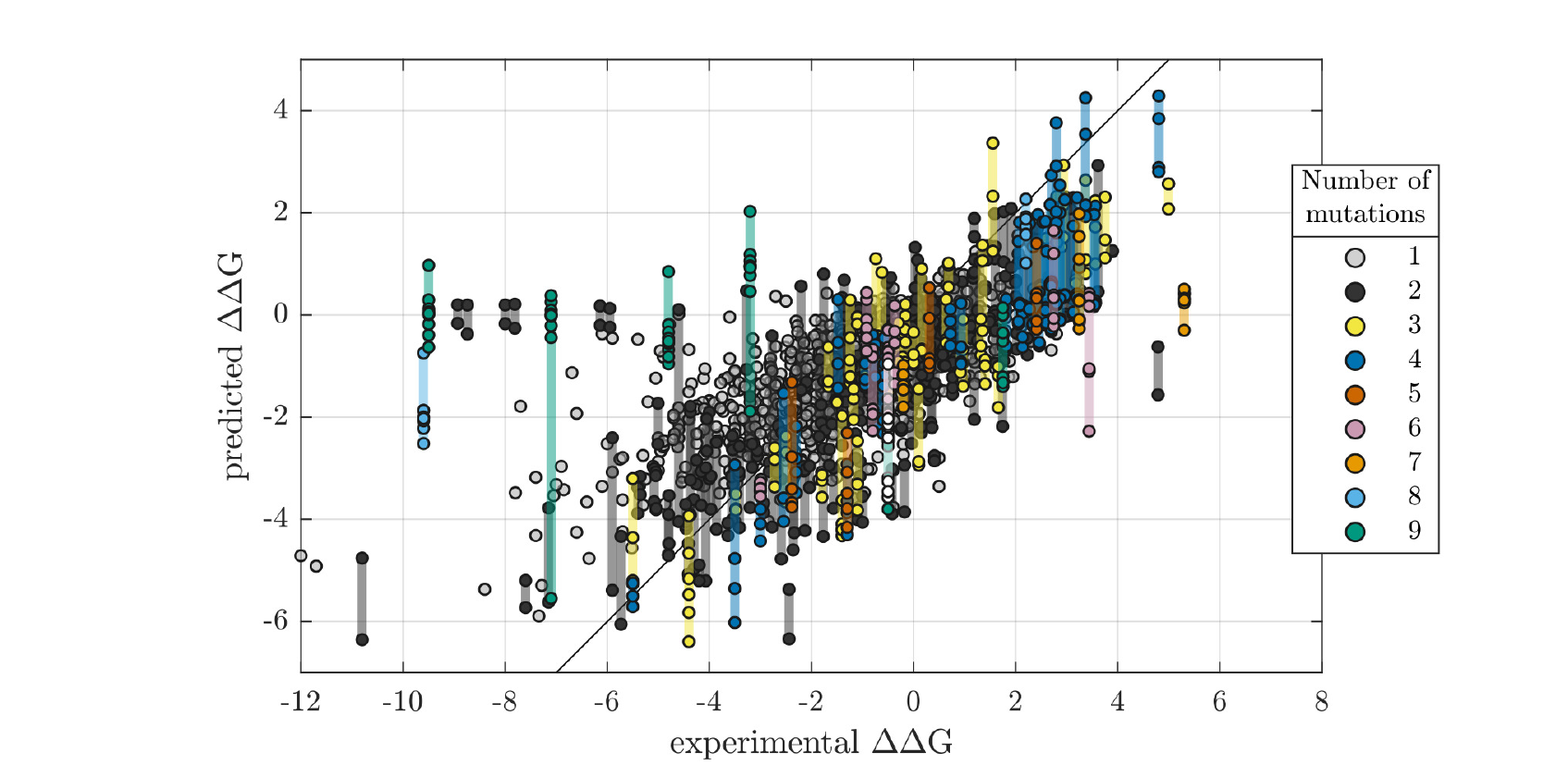}}
\caption{Position-level predictions for all 15 proteins. When the effects of a mutant are predicted multiple times, they are connected by a line.}\label{fig:pos}
\end{figure}

\clearpage

{\begin{table}[!htb]
\caption{(Continues on the next page)Comparison of different methods on the $15$ protein dataset with respect to $\rho$ and $\rmse$. Off-the-shelf implementations of Rosetta, mCSM and PoPMuSiC are used directly without cross-validation. \label{tab:results}
}
{\resizebox{\textwidth}{!}{
\begin{tabular}
{@{}cl | rcl | rcl | rcl || rcl | rcl | rcl @{}}\toprule 
 &  & \multicolumn{9}{c||}{Correlation $\rho$} & \multicolumn{9}{c}{$\rmse$}\\
 &  & \multicolumn{3}{c|}{Point mutations} & \multicolumn{3}{c|}{Multiple mutations} & \multicolumn{3}{c||}{All mutations} & \multicolumn{3}{c|}{Point mutations} & \multicolumn{3}{c|}{Multiple mutations} & \multicolumn{3}{c}{All mutations} \\
 \cline{3-20}
 & & \multicolumn{3}{c|}{cross-validation level} & \multicolumn{3}{c|}{cross-validation level} & \multicolumn{3}{c||}{cross-validation level} & \multicolumn{3}{c|}{cross-validation level} & \multicolumn{3}{c|}{cross-validation level} & \multicolumn{3}{c}{cross-validation level} \\
Protein & Method & \:\, mut. & pos. & \multicolumn{1}{l|}{prot.} & \:\, mut. & pos. & \multicolumn{1}{l|}{prot.} & \:\, mut. &  pos. &  prot. & \:\, mut. & pos. & \multicolumn{1}{l|}{prot.} & \:\, mut. & pos. & \multicolumn{1}{l|}{prot.} & \:\, mut. & pos. &  prot.\\
\hline
 \multirow{6}{*}{2LZM}&mGPfusion & 0.87 & 0.80 & 0.75 & 0.96 & 0.83 & 0.64 & 0.90 & 0.76 & 0.68 & 0.82 & 1.02 & 1.12 & 0.57 & 1.53 & 2.14 & 0.76 & 1.30 & 1.43 \\
&mGPfusion, only B62 & 0.86 & 0.77 & 0.75 & 0.96 & 0.87 & 0.64 & 0.90 & 0.82 & 0.69 &0.84 & 1.08 & 1.12 & 0.59 & 1.16 & 2.11 & 0.79 & 1.13 & 1.42 \\
&mGP & 0.86 & 0.59 & \:\:\:\:- & 0.97 & 0.85 & \:\:\:\:- & 0.90 & 0.72 & \:\:\:\:- &0.82 & 1.34 & \:\:\:\:- & 0.48 & 1.24 & \:\:\:\:- & 0.75 & 1.29 & \:\:\:\:- \\
&mGP, only B62 & 0.75 & 0.37 & \:\:\:\:- & 0.94 & 0.77 & \:\:\:\:- & 0.82 & 0.61 & \:\:\:\:- &1.12 & 1.93 & \:\:\:\:- & 0.78 & 1.44 & \:\:\:\:- & 1.05 & 1.70 & \:\:\:\:- \\
&Rosetta scaled & 0.74 & 0.73 & \:\:\:\:- & 0.68 & 0.66 & \:\:\:\:- & 0.70 & 0.65 & \:\:\:\:- &1.05 & 1.06 & \:\:\:\:- & 1.65 & 1.84 & \:\:\:\:- & 1.23 & 1.51 & \:\:\:\:- \\ \cline{2-20}
&Rosetta & \multicolumn{3}{c|}{0.75} &\multicolumn{3}{c|}{0.68} & \multicolumn{3}{c||}{0.71} & \multicolumn{3}{c|}{1.13} & \multicolumn{3}{c|}{1.93} & \multicolumn{3}{c}{1.37} \\
	                    &mCSM               	&\multicolumn{3}{c|}{0.57} &\multicolumn{3}{c|}{-}    &\multicolumn{3}{c||}{-}&\multicolumn{3}{c|}{1.27}&\multicolumn{3}{c|}{-}	&\multicolumn{3}{c}{-}	\\
                    	&PoPMuSiC           	&\multicolumn{3}{c|}{0.71} &\multicolumn{3}{c|}{-}    &\multicolumn{3}{c||}{-}&\multicolumn{3}{c|}{1.11}&\multicolumn{3}{c|}{-}	&\multicolumn{3}{c}{-}	\\\hline
\multirow{6}{*}{1BNI}	&mGPfusion & 0.77 & 0.64 & 0.62 & 0.86 & 0.70 & 0.39 & 0.77 & 0.55 & 0.57 & 1.21 & 1.37 & 1.69 & 1.28 & 2.49 & 2.17 & 1.22 & 1.67 & 1.75 \\
&mGPfusion, only B62 & 0.74 & 0.61 & 0.62 & 0.84 & 0.79 & 0.40 & 0.74 & 0.60 & 0.57 &1.27 & 1.45 & 1.69 & 1.24 & 2.03 & 2.14 & 1.27 & 1.60 & 1.75 \\
&mGP & 0.81 & 0.65 & \:\:\:\:- & 0.86 & 0.82 & \:\:\:\:- & 0.80 & 0.63 & \:\:\:\:- &1.08 & 1.40 & \:\:\:\:- & 1.32 & 2.00 & \:\:\:\:- & 1.11 & 1.55 & \:\:\:\:- \\
&mGP, only B62 & 0.61 & 0.48 & \:\:\:\:- & 0.87 & 0.85 & \:\:\:\:- & 0.63 & 0.41 & \:\:\:\:- &1.61 & 2.32 & \:\:\:\:- & 1.15 & 1.67 & \:\:\:\:- & 1.57 & 2.20 & \:\:\:\:- \\
&Rosetta scaled & 0.59 & 0.58 & \:\:\:\:- & 0.17 & 0.29 & \:\:\:\:- & 0.53 & 0.45 & \:\:\:\:- &1.58 & 1.58 & \:\:\:\:- & 2.51 & 2.90 & \:\:\:\:- & 1.70 & 1.94 & \:\:\:\:- \\ \cline{2-20}
&Rosetta & \multicolumn{3}{c|}{0.62} &\multicolumn{3}{c|}{0.18} & \multicolumn{3}{c||}{0.56} & \multicolumn{3}{c|}{1.70} & \multicolumn{3}{c|}{2.33} & \multicolumn{3}{c}{1.77} \\
	                    &mCSM	                &\multicolumn{3}{c|}{0.60} &\multicolumn{3}{c|}{-}	&\multicolumn{3}{c||}{-}	&\multicolumn{3}{c|}{1.62} &\multicolumn{3}{c|}{-}	&\multicolumn{3}{c}{-}	\\
	                    &PoPMuSiC           	&\multicolumn{3}{c|}{0.66} &\multicolumn{3}{c|}{-}	&\multicolumn{3}{c||}{-}	&\multicolumn{3}{c|}{1.53} &\multicolumn{3}{c|}{-}	&\multicolumn{3}{c}{-}	\\
\hline
\multirow{6}{*}{1VQB}	&mGPfusion & 0.67 & 0.50 & 0.49 & 0.93 & 0.83 & 0.75 & 0.76 & 0.69 & 0.60 & 1.71 & 1.94 & 2.25 & 1.15 & 1.62 & 2.06 & 1.59 & 1.82 & 2.20 \\
&mGPfusion, only B62 & 0.65 & 0.53 & 0.50 & 0.91 & 0.82 & 0.73 & 0.75 & 0.69 & 0.58 &1.75 & 1.94 & 2.25 & 1.35 & 1.82 & 2.43 & 1.66 & 1.89 & 2.30 \\
&mGP & 0.79 & 0.12 & \:\:\:\:- & 0.96 & 0.70 & \:\:\:\:- & 0.85 & 0.50 & \:\:\:\:- &1.41 & 2.41 & \:\:\:\:- & 0.70 & 1.97 & \:\:\:\:- & 1.27 & 2.24 & \:\:\:\:- \\
&mGP, only B62 & 0.79 & 0.29 & \:\:\:\:- & 0.97 & 0.75 & \:\:\:\:- & 0.85 & 0.55 & \:\:\:\:- &1.50 & 2.89 & \:\:\:\:- & 0.63 & 2.30 & \:\:\:\:- & 1.33 & 2.66 & \:\:\:\:- \\
&Rosetta scaled & 0.47 & 0.46 & \:\:\:\:- & 0.71 & 0.68 & \:\:\:\:- & 0.57 & 0.59 & \:\:\:\:- &1.99 & 2.00 & \:\:\:\:- & 1.90 & 1.96 & \:\:\:\:- & 1.97 & 1.99 & \:\:\:\:- \\ \cline{2-20}
&Rosetta & \multicolumn{3}{c|}{0.49} &\multicolumn{3}{c|}{0.73} & \multicolumn{3}{c||}{0.59} & \multicolumn{3}{c|}{2.26} & \multicolumn{3}{c|}{2.06} & \multicolumn{3}{c}{2.21} \\
	&mCSM	&\multicolumn{3}{c|}{0.53} &\multicolumn{3}{c|}{-}	&\multicolumn{3}{c||}{-}	&\multicolumn{3}{c|}{2.24} &\multicolumn{3}{c|}{-}	&\multicolumn{3}{c}{-}	\\
	&PoPMuSiC	&\multicolumn{3}{c|}{0.51} &\multicolumn{3}{c|}{-}	&\multicolumn{3}{c||}{-}	&\multicolumn{3}{c|}{2.29} &\multicolumn{3}{c|}{-}	&\multicolumn{3}{c}{-}	\\
\hline
\multirow{6}{*}{1LZ1}	&mGPfusion & 0.75 & 0.59 & 0.58 & 1.00 & 0.11 & -1.00 & 0.77 & 0.57 & 0.52 & 0.83 & 0.99 & 1.06 & 0.56 & 2.40 & 3.75 & 0.83 & 1.07 & 1.16 \\
&mGPfusion, only B62 & 0.73 & 0.56 & 0.59 & 1.00 & 0.05 & -1.00 & 0.76 & 0.56 & 0.53 &0.86 & 1.03 & 1.06 & 1.10 & 2.41 & 3.51 & 0.87 & 1.10 & 1.14 \\
&mGP & 0.75 & 0.39 & \:\:\:\:- & 1.00 & 0.56 & \:\:\:\:- & 0.78 & 0.47 & \:\:\:\:- &0.81 & 1.15 & \:\:\:\:- & 0.13 & 1.65 & \:\:\:\:- & 0.80 & 1.17 & \:\:\:\:- \\
&mGP, only B62 & 0.71 & -0.31 & \:\:\:\:- & 1.00 & 0.42 & \:\:\:\:- & 0.74 & 0.21 & \:\:\:\:- &0.91 & 1.43 & \:\:\:\:- & 0.27 & 2.36 & \:\:\:\:- & 0.90 & 1.47 & \:\:\:\:- \\
&Rosetta scaled & 0.57 & 0.55 & \:\:\:\:- & -1.00 & -1.00 & \:\:\:\:- & 0.53 & 0.46 & \:\:\:\:- &0.99 & 1.01 & \:\:\:\:- & 3.23 & 3.25 & \:\:\:\:- & 1.07 & 1.15 & \:\:\:\:- \\ \cline{2-20}
&Rosetta & \multicolumn{3}{c|}{0.59} &\multicolumn{3}{c|}{-1.00} & \multicolumn{3}{c||}{0.55} & \multicolumn{3}{c|}{1.04} & \multicolumn{3}{c|}{3.41} & \multicolumn{3}{c}{1.12} \\
	&mCSM	&\multicolumn{3}{c|}{0.67} &\multicolumn{3}{c|}{-}	&\multicolumn{3}{c||}{-}	&\multicolumn{3}{c|}{0.97} &\multicolumn{3}{c|}{-}	&\multicolumn{3}{c}{-}	\\
	&PoPMuSiC	&\multicolumn{3}{c|}{0.64} &\multicolumn{3}{c|}{-}	&\multicolumn{3}{c||}{-}	&\multicolumn{3}{c|}{0.95} &\multicolumn{3}{c|}{-}	&\multicolumn{3}{c}{-}	\\
\hline
\multirow{6}{*}{2CI2}	&mGPfusion & 0.73 & 0.72 & 0.64 & 0.95 & 0.87 & 0.85 & 0.82 & 0.81 & 0.71 & 0.85 & 0.90 & 1.07 & 0.55 & 0.80 & 1.21 & 0.80 & 0.86 & 1.10 \\
&mGPfusion, only B62 & 0.69 & 0.67 & 0.63 & 0.92 & 0.86 & 0.86 & 0.79 & 0.79 & 0.72 &0.91 & 0.97 & 1.07 & 0.71 & 1.01 & 1.12 & 0.87 & 0.99 & 1.08 \\
&mGP & 0.65 & 0.61 & \:\:\:\:- & 0.92 & 0.79 & \:\:\:\:- & 0.76 & 0.72 & \:\:\:\:- &0.95 & 1.02 & \:\:\:\:- & 0.66 & 1.01 & \:\:\:\:- & 0.90 & 1.02 & \:\:\:\:- \\
&mGP, only B62 & 0.51 & 0.74 & \:\:\:\:- & 0.92 & 0.71 & \:\:\:\:- & 0.68 & 0.63 & \:\:\:\:- &1.16 & 1.39 & \:\:\:\:- & 0.71 & 1.40 & \:\:\:\:- & 1.08 & 1.39 & \:\:\:\:- \\
&Rosetta scaled & 0.60 & 0.60 & \:\:\:\:- & 0.61 & 0.61 & \:\:\:\:- & 0.63 & 0.63 & \:\:\:\:- &1.00 & 1.00 & \:\:\:\:- & 1.27 & 1.27 & \:\:\:\:- & 1.06 & 1.11 & \:\:\:\:- \\ \cline{2-20}
&Rosetta & \multicolumn{3}{c|}{0.63} &\multicolumn{3}{c|}{0.62} & \multicolumn{3}{c||}{0.65} & \multicolumn{3}{c|}{1.09} & \multicolumn{3}{c|}{1.30} & \multicolumn{3}{c}{1.13} \\
	&mCSM	&\multicolumn{3}{c|}{0.74} &\multicolumn{3}{c|}{-}	&\multicolumn{3}{c||}{-}	&\multicolumn{3}{c|}{0.86} &\multicolumn{3}{c|}{-}	&\multicolumn{3}{c}{-}	\\
	&PoPMuSiC	&\multicolumn{3}{c|}{0.75} &\multicolumn{3}{c|}{-}	&\multicolumn{3}{c||}{-}	&\multicolumn{3}{c|}{0.85} &\multicolumn{3}{c|}{-}	&\multicolumn{3}{c}{-}	\\
\hline
\multirow{6}{*}{1PGA}	&mGPfusion & 0.68 & 0.47 & 0.69 & 0.90 & 0.35 & 0.32 & 0.85 & 0.43 & 0.50 & 1.26 & 1.54 & 1.64 & 0.53 & 2.09 & 2.74 & 0.88 & 2.00 & 2.38 \\
&mGPfusion, only B62 & 0.82 & 0.59 & 0.71 & 0.76 & 0.60 & 0.62 & 0.81 & 0.53 & 0.70 &0.87 & 1.22 & 1.06 & 0.69 & 0.95 & 0.88 & 0.83 & 1.13 & 1.02 \\
&mGP & 0.62 & 0.61 & \:\:\:\:- & 0.93 & -0.24 & \:\:\:\:- & 0.84 & -0.14 & \:\:\:\:- &1.40 & 1.58 & \:\:\:\:- & 0.45 & 3.01 & \:\:\:\:- & 0.94 & 2.81 & \:\:\:\:- \\
&mGP, only B62 & 0.57 & -0.46 & \:\:\:\:- & 0.92 & -0.08 & \:\:\:\:- & 0.81 & 0.06 & \:\:\:\:- &1.53 & 1.73 & \:\:\:\:- & 0.47 & 2.07 & \:\:\:\:- & 1.02 & 2.01 & \:\:\:\:- \\
&Rosetta scaled & 0.69 & 0.59 & \:\:\:\:- & 0.09 & 0.07 & \:\:\:\:- & 0.24 & 0.11 & \:\:\:\:- &1.21 & 1.42 & \:\:\:\:- & 2.81 & 3.09 & \:\:\:\:- & 2.33 & 2.87 & \:\:\:\:- \\ \cline{2-20}
&Rosetta & \multicolumn{3}{c|}{0.69} &\multicolumn{3}{c|}{0.03} & \multicolumn{3}{c||}{0.28} & \multicolumn{3}{c|}{1.70} & \multicolumn{3}{c|}{3.51} & \multicolumn{3}{c}{2.95} \\
	&mCSM	&\multicolumn{3}{c|}{-0.10} &\multicolumn{3}{c|}{-}	&\multicolumn{3}{c||}{-}	&\multicolumn{3}{c|}{1.94} &\multicolumn{3}{c|}{-}	&\multicolumn{3}{c}{-}	\\
	&PoPMuSiC	&\multicolumn{3}{c|}{0.28} &\multicolumn{3}{c|}{-}	&\multicolumn{3}{c||}{-}	&\multicolumn{3}{c|}{1.89} &\multicolumn{3}{c|}{-}	&\multicolumn{3}{c}{-}	\\
\hline
\multirow{6}{*}{2RN2}	&mGPfusion & 0.79 & 0.58 & 0.71 & 0.78 & 0.60 & 0.61 & 0.79 & 0.53 & 0.70 & 0.91 & 1.21 & 1.05 & 0.67 & 1.01 & 0.91 & 0.86 & 1.14 & 1.02 \\
&mGPfusion, only B62 & 0.82 & 0.59 & 0.71 & 0.76 & 0.60 & 0.62 & 0.81 & 0.53 & 0.70 &0.87 & 1.22 & 1.59 & 0.69 & 0.95 & 1.18 & 0.83 & 1.13 & 1.51 \\
&mGP & 0.77 & 0.12 & \:\:\:\:- & 0.75 & 0.42 & \:\:\:\:- & 0.77 & 0.22 & \:\:\:\:- &0.93 & 1.45 & \:\:\:\:- & 0.74 & 1.21 & \:\:\:\:- & 0.89 & 1.36 & \:\:\:\:- \\
&mGP, only B62 & 0.83 & 0.09 & \:\:\:\:- & 0.77 & 0.42 & \:\:\:\:- & 0.82 & 0.23 & \:\:\:\:- &0.82 & 1.45 & \:\:\:\:- & 0.68 & 1.20 & \:\:\:\:- & 0.80 & 1.36 & \:\:\:\:- \\
&Rosetta scaled & 0.66 & 0.64 & \:\:\:\:- & 0.48 & 0.50 & \:\:\:\:- & 0.62 & 0.57 & \:\:\:\:- &1.09 & 1.13 & \:\:\:\:- & 1.20 & 1.08 & \:\:\:\:- & 1.12 & 1.11 & \:\:\:\:- \\ \cline{2-20}
&Rosetta & \multicolumn{3}{c|}{0.70} &\multicolumn{3}{c|}{0.47} & \multicolumn{3}{c||}{0.65} & \multicolumn{3}{c|}{1.07} & \multicolumn{3}{c|}{1.25} & \multicolumn{3}{c}{1.11} \\
	&mCSM	&\multicolumn{3}{c|}{0.71} &\multicolumn{3}{c|}{-}	&\multicolumn{3}{c||}{-}	&\multicolumn{3}{c|}{1.04} &\multicolumn{3}{c|}{-}	&\multicolumn{3}{c}{-}	\\
	&PoPMuSiC	&\multicolumn{3}{c|}{0.71} &\multicolumn{3}{c|}{-}	&\multicolumn{3}{c||}{-}	&\multicolumn{3}{c|}{1.16} &\multicolumn{3}{c|}{-}	&\multicolumn{3}{c}{-}	\\
\hline
\multirow{6}{*}{1CSP}	&mGPfusion & 0.85 & 0.23 & 0.33 & 0.92 & 0.73 & 0.48 & 0.92 & 0.75 & 0.38 & 0.64 & 1.04 & 1.10 & 0.91 & 1.66 & 2.58 & 0.75 & 1.45 & 1.80 \\
&mGPfusion, only B62 & 0.86 & 0.22 & 0.34 & 0.91 & 0.69 & 0.59 & 0.91 & 0.72 & 0.54 &0.65 & 1.04 & 1.10 & 0.96 & 1.87 & 2.13 & 0.78 & 1.60 & 1.57 \\
&mGP & 0.88 & -0.06 & \:\:\:\:- & 0.94 & 0.75 & \:\:\:\:- & 0.94 & 0.77 & \:\:\:\:- &0.54 & 1.07 & \:\:\:\:- & 0.76 & 1.51 & \:\:\:\:- & 0.63 & 1.36 & \:\:\:\:- \\
&mGP, only B62 & 0.87 & -0.37 & \:\:\:\:- & 0.92 & 0.71 & \:\:\:\:- & 0.92 & 0.72 & \:\:\:\:- &0.60 & 1.12 & \:\:\:\:- & 0.86 & 1.59 & \:\:\:\:- & 0.71 & 1.43 & \:\:\:\:- \\
&Rosetta scaled & 0.23 & 0.20 & \:\:\:\:- & 0.68 & 0.69 & \:\:\:\:- & 0.59 & 0.64 & \:\:\:\:- &1.04 & 1.06 & \:\:\:\:- & 2.19 & 2.29 & \:\:\:\:- & 1.58 & 1.92 & \:\:\:\:- \\ \cline{2-20}
&Rosetta & \multicolumn{3}{c|}{0.33} &\multicolumn{3}{c|}{0.68} & \multicolumn{3}{c||}{0.60} & \multicolumn{3}{c|}{1.11} & \multicolumn{3}{c|}{1.92} & \multicolumn{3}{c}{1.47} \\
	&mCSM	&\multicolumn{3}{c|}{0.42} &\multicolumn{3}{c|}{-}	&\multicolumn{3}{c||}{-}	&\multicolumn{3}{c|}{1.02} &\multicolumn{3}{c|}{-}	&\multicolumn{3}{c}{-}	\\
	&PoPMuSiC	&\multicolumn{3}{c|}{0.48} &\multicolumn{3}{c|}{-}	&\multicolumn{3}{c||}{-}	&\multicolumn{3}{c|}{0.99} &\multicolumn{3}{c|}{-}	&\multicolumn{3}{c}{-}	\\
\bottomrule
\end{tabular}}}{}
\end{table}}
\addtocounter{table}{-1}

{\begin{table}[!htb]
\caption{(Continued) Comparison of different methods on the $15$ protein dataset with respect to $\rho$ and $\rmse$. Off-the-shelf implementations of Rosetta, mCSM and PoPMuSiC are used directly without cross-validation. \label{tab:results2}
}
{\resizebox{\textwidth}{!}{
\begin{tabular}
{@{}cl | rcl | rcl | rcl || rcl | rcl | rcl @{}}\toprule 
 &  & \multicolumn{9}{c||}{Correlation $\rho$} & \multicolumn{9}{c}{$\rmse$}\\
 &  & \multicolumn{3}{c|}{Point mutations} & \multicolumn{3}{c|}{Multiple mutations} & \multicolumn{3}{c||}{All mutations} & \multicolumn{3}{c|}{Point mutations} & \multicolumn{3}{c|}{Multiple mutations} & \multicolumn{3}{c}{All mutations} \\
 \cline{3-20}
 & & \multicolumn{3}{c|}{cross-validation level} & \multicolumn{3}{c|}{cross-validation level} & \multicolumn{3}{c||}{cross-validation level} & \multicolumn{3}{c|}{cross-validation level} & \multicolumn{3}{c|}{cross-validation level} & \multicolumn{3}{c}{cross-validation level} \\
Protein & Method & \:\, mut. & pos. & \multicolumn{1}{l|}{prot.} & \:\, mut. & pos. & \multicolumn{1}{l|}{prot.} & \:\, mut. &  pos. &  prot. & \:\, mut. & pos. & \multicolumn{1}{l|}{prot.} & \:\, mut. & pos. & \multicolumn{1}{l|}{prot.} & \:\, mut. & pos. &  prot.\\
\hline
\multirow{6}{*}{1BVC}	&mGPfusion & 0.41 & 0.43 & 0.48 & -0.28 & -0.25 & -0.63 & 0.08 & -0.09 & 0.09 & 0.74 & 0.72 & 1.65 & 1.64 & 1.48 & 2.70 & 1.09 & 1.21 & 2.02 \\
&mGPfusion, only B62 & 0.50 & 0.48 & 0.48 & -0.25 & -0.25 & -0.66 & 0.14 & -0.07 & 0.14 &0.70 & 0.71 & 1.65 & 1.62 & 1.47 & 2.43 & 1.06 & 1.20 & 1.92 \\
&mGP & -0.05 & -0.12 & \:\:\:\:- & -0.10 & -0.21 & \:\:\:\:- & -0.13 & -0.23 & \:\:\:\:- &0.99 & 1.00 & \:\:\:\:- & 1.30 & 1.35 & \:\:\:\:- & 1.09 & 1.21 & \:\:\:\:- \\
&mGP, only B62 & -0.05 & 0.06 & \:\:\:\:- & -0.11 & -0.21 & \:\:\:\:- & -0.14 & -0.20 & \:\:\:\:- &0.99 & 0.99 & \:\:\:\:- & 1.29 & 1.33 & \:\:\:\:- & 1.09 & 1.20 & \:\:\:\:- \\
&Rosetta scaled & 0.42 & 0.40 & \:\:\:\:- & -0.63 & -0.57 & \:\:\:\:- & 0.09 & -0.09 & \:\:\:\:- &0.75 & 0.76 & \:\:\:\:- & 1.65 & 1.45 & \:\:\:\:- & 1.10 & 1.20 & \:\:\:\:- \\ \cline{2-20}
&Rosetta & \multicolumn{3}{c|}{0.47} &\multicolumn{3}{c|}{-0.65} & \multicolumn{3}{c||}{0.14} & \multicolumn{3}{c|}{1.67} & \multicolumn{3}{c|}{2.45} & \multicolumn{3}{c}{1.94} \\
	&mCSM	&\multicolumn{3}{c|}{0.47} &\multicolumn{3}{c|}{-}	&\multicolumn{3}{c||}{-}	&\multicolumn{3}{c|}{1.00} &\multicolumn{3}{c|}{-}	&\multicolumn{3}{c}{-}	\\
	&PoPMuSiC	&\multicolumn{3}{c|}{0.60} &\multicolumn{3}{c|}{-}	&\multicolumn{3}{c||}{-}	&\multicolumn{3}{c|}{0.85} &\multicolumn{3}{c|}{-}	&\multicolumn{3}{c}{-}	\\
\hline
\multirow{6}{*}{4LYZ}	&mGPfusion & 0.61 & 0.27 & 0.34 & 0.95 & 0.57 & 0.66 & 0.65 & 0.34 & 0.35 & 1.47 & 1.65 & 3.46 & 0.69 & 1.02 & 2.35 & 1.35 & 1.45 & 3.26 \\
&mGPfusion, only B62 & 0.59 & 0.30 & 0.34 & 0.96 & 0.64 & 0.64 & 0.64 & 0.40 & 0.35 &1.48 & 1.64 & 3.50 & 0.57 & 0.93 & 2.17 & 1.35 & 1.43 & 3.27 \\
&mGP & 0.65 & -0.05 & \:\:\:\:- & 0.96 & 0.46 & \:\:\:\:- & 0.68 & 0.18 & \:\:\:\:- &1.38 & 1.78 & \:\:\:\:- & 0.36 & 1.15 & \:\:\:\:- & 1.24 & 1.59 & \:\:\:\:- \\
&mGP, only B62 & 0.62 & 0.27 & \:\:\:\:- & 0.97 & 0.47 & \:\:\:\:- & 0.64 & 0.21 & \:\:\:\:- &1.48 & 1.84 & \:\:\:\:- & 0.31 & 1.16 & \:\:\:\:- & 1.33 & 1.63 & \:\:\:\:- \\
&Rosetta scaled & 0.29 & 0.28 & \:\:\:\:- & 0.70 & 0.68 & \:\:\:\:- & 0.33 & 0.34 & \:\:\:\:- &2.26 & 2.22 & \:\:\:\:- & 1.22 & 1.29 & \:\:\:\:- & 2.09 & 1.94 & \:\:\:\:- \\ \cline{2-20}
&Rosetta & \multicolumn{3}{c|}{0.33} &\multicolumn{3}{c|}{0.71} & \multicolumn{3}{c||}{0.35} & \multicolumn{3}{c|}{3.61} & \multicolumn{3}{c|}{2.05} & \multicolumn{3}{c}{3.35} \\
	&mCSM	&\multicolumn{3}{c|}{0.55} &\multicolumn{3}{c|}{-}	&\multicolumn{3}{c||}{-}	&\multicolumn{3}{c|}{1.43} &\multicolumn{3}{c|}{-}	&\multicolumn{3}{c}{-}	\\
	&PoPMuSiC	&\multicolumn{3}{c|}{0.59} &\multicolumn{3}{c|}{-}	&\multicolumn{3}{c||}{-}	&\multicolumn{3}{c|}{1.45} &\multicolumn{3}{c|}{-}	&\multicolumn{3}{c}{-}	\\
\hline
\multirow{6}{*}{1RTB}	&mGPfusion & 0.92 & 0.81 & 0.70 & 0.78 & 0.74 & 0.75 & 0.86 & 0.71 & 0.68 & 1.25 & 1.73 & 2.44 & 2.22 & 2.24 & 1.71 & 1.40 & 1.85 & 2.36 \\
&mGPfusion, only B62 & 0.89 & 0.79 & 0.70 & 0.73 & 0.73 & 0.67 & 0.86 & 0.75 & 0.69 &1.48 & 1.85 & 2.44 & 1.82 & 1.81 & 1.90 & 1.53 & 1.84 & 2.38 \\
&mGP & 0.92 & 0.69 & \:\:\:\:- & 0.78 & 0.30 & \:\:\:\:- & 0.76 & 0.18 & \:\:\:\:- &1.26 & 2.62 & \:\:\:\:- & 3.44 & 3.39 & \:\:\:\:- & 1.69 & 2.81 & \:\:\:\:- \\
&mGP, only B62 & 0.91 & 0.58 & \:\:\:\:- & 0.48 & 0.58 & \:\:\:\:- & 0.85 & 0.16 & \:\:\:\:- &1.67 & 3.37 & \:\:\:\:- & 2.12 & 2.11 & \:\:\:\:- & 1.73 & 3.14 & \:\:\:\:- \\
&Rosetta scaled & 0.65 & 0.61 & \:\:\:\:- & 0.73 & 0.73 & \:\:\:\:- & 0.67 & 0.65 & \:\:\:\:- &1.99 & 2.08 & \:\:\:\:- & 1.63 & 1.62 & \:\:\:\:- & 1.95 & 1.99 & \:\:\:\:- \\ \cline{2-20}
&Rosetta & \multicolumn{3}{c|}{0.69} &\multicolumn{3}{c|}{0.73} & \multicolumn{3}{c||}{0.70} & \multicolumn{3}{c|}{2.45} & \multicolumn{3}{c|}{2.01} & \multicolumn{3}{c}{2.40} \\
	&mCSM	&\multicolumn{3}{c|}{0.68} &\multicolumn{3}{c|}{-}	&\multicolumn{3}{c||}{-}	&\multicolumn{3}{c|}{2.33} &\multicolumn{3}{c|}{-}	&\multicolumn{3}{c}{-}	\\
	&PoPMuSiC	&\multicolumn{3}{c|}{0.72} &\multicolumn{3}{c|}{-}	&\multicolumn{3}{c||}{-}	&\multicolumn{3}{c|}{2.22} &\multicolumn{3}{c|}{-}	&\multicolumn{3}{c}{-}	\\
\hline
\multirow{6}{*}{1PIN$^*$}&mGPfusion & 0.64 & 0.49 & 0.53 & - & - & \:\:\:\:- & 0.64 & 0.49 & 0.53 & 0.47 & 0.53 & 1.00 & - & - & \:\:\:\:- & 0.47 & 0.53 & 1.00 \\
&mGPfusion, only B62 & 0.59 & 0.51 & 0.53 & \:\:\:\:- & \:\:\:\:- & \:\:\:\:- & 0.59 & 0.51 & 0.53 & 0.50 & 0.54 & 1.00 & - & - & \:\:\:\:- & 0.50 & 0.54 & 1.00 \\
&mGP & 0.61 & 0.40 & \:\:\:\:- & - & - & \:\:\:\:- & 0.61 & 0.40 & \:\:\:\:- &0.49 & 0.58 & \:\:\:\:- & - & - & \:\:\:\:- & 0.49 & 0.58 & \:\:\:\:- \\
&mGP, only B62 & 0.23 & 0.25 & \:\:\:\:- & \:\:\:\:- & \:\:\:\:- & \:\:\:\:- & 0.23 & 0.25 & \:\:\:\:- &0.70 & 0.76 & \:\:\:\:- & - & - & \:\:\:\:- & 0.70 & 0.76 & \:\:\:\:- \\
&Rosetta scaled & 0.50 & 0.49 & \:\:\:\:- & \:\:\:\:- & \:\:\:\:- & \:\:\:\:- & 0.50 & 0.49 & \:\:\:\:- &0.54 & 0.54 & \:\:\:\:- & - & - & \:\:\:\:- & 0.54 & 0.54 & \:\:\:\:- \\ \cline{2-20}
&Rosetta & \multicolumn{3}{c|}{0.53} &\multicolumn{3}{c|}{-} & \multicolumn{3}{c||}{0.53} & \multicolumn{3}{c|}{0.99} & \multicolumn{3}{c|}{} & \multicolumn{3}{c}{0.99} \\
	                &mCSM	                &\multicolumn{3}{c|}{0.72} &\multicolumn{3}{c|}{-} &\multicolumn{3}{c||}{0.72} &\multicolumn{3}{c|}{0.59} &\multicolumn{3}{c|}{-} &\multicolumn{3}{c}{0.59} \\
	                &PoPMuSiC	            &\multicolumn{3}{c|}{0.60} &\multicolumn{3}{c|}{-} &\multicolumn{3}{c||}{0.60} &\multicolumn{3}{c|}{0.70} &\multicolumn{3}{c|}{-} &\multicolumn{3}{c}{0.70} \\
\hline
\multirow{6}{*}{1RN1}&mGPfusion & 0.83 & 0.62 & 0.67 & 0.97 & 0.38 & -0.19 & 0.84 & 0.55 & 0.65 & 0.85 & 1.17 & 1.18 & 0.19 & 0.91 & 1.28 & 0.81 & 1.13 & 1.19 \\
&mGPfusion, only B62 & 0.81 & 0.58 & 0.67 & 0.96 & 0.42 & -0.01 & 0.81 & 0.51 & 0.66 &0.92 & 1.24 & 1.18 & 0.35 & 0.77 & 0.99 & 0.88 & 1.17 & 1.17 \\
&mGP & 0.74 & 0.04 & \:\:\:\:- & 0.99 & 0.62 & \:\:\:\:- & 0.75 & 0.04 & \:\:\:\:- &1.06 & 1.75 & \:\:\:\:- & 0.12 & 0.58 & \:\:\:\:- & 1.01 & 1.61 & \:\:\:\:- \\
&mGP, only B62 & 0.77 & 0.27 & \:\:\:\:- & 0.99 & 0.50 & \:\:\:\:- & 0.78 & 0.10 & \:\:\:\:- &1.00 & 1.78 & \:\:\:\:- & 0.13 & 0.82 & \:\:\:\:- & 0.95 & 1.66 & \:\:\:\:- \\
&Rosetta scaled & 0.63 & 0.57 & \:\:\:\:- & 0.21 & 0.18 & \:\:\:\:- & 0.61 & 0.54 & \:\:\:\:- &1.12 & 1.18 & \:\:\:\:- & 0.98 & 1.01 & \:\:\:\:- & 1.10 & 1.15 & \:\:\:\:- \\ \cline{2-20}
&Rosetta & \multicolumn{3}{c|}{0.67} &\multicolumn{3}{c|}{0.20} & \multicolumn{3}{c||}{0.65} & \multicolumn{3}{c|}{1.20} & \multicolumn{3}{c|}{1.24} & \multicolumn{3}{c}{1.20} \\
	&mCSM	    &\multicolumn{3}{c|}{0.76} &\multicolumn{3}{c|}{-}	&\multicolumn{3}{c||}{-}	&\multicolumn{3}{c|}{0.97} &\multicolumn{3}{c|}{-}	&\multicolumn{3}{c}{-}	\\
	&PoPMuSiC	&\multicolumn{3}{c|}{0.62} &\multicolumn{3}{c|}{-}	&\multicolumn{3}{c||}{-}	&\multicolumn{3}{c|}{1.14} &\multicolumn{3}{c|}{-}	&\multicolumn{3}{c}{-}	\\
\hline
\multirow{6}{*}{1RGG}	&mGPfusion & 0.69 & 0.51 & 0.61 & 0.42 & 0.66 & 0.96 & 0.52 & 0.52 & 0.73 & 1.41 & 1.70 & 1.56 & 5.67 & 5.95 & 4.20 & 2.65 & 4.75 & 2.23 \\
&mGPfusion, only B62 & 0.69 & 0.52 & 0.61 & 0.22 & 0.30 & 0.93 & 0.40 & 0.33 & 0.55 &1.42 & 1.68 & 1.56 & 6.21 & 6.14 & 5.43 & 2.85 & 4.90 & 2.64 \\
&mGP & 0.71 & -0.61 & \:\:\:\:- & 0.19 & 0.08 & \:\:\:\:- & 0.35 & -0.05 & \:\:\:\:- &1.38 & 2.24 & \:\:\:\:- & 6.81 & 6.88 & \:\:\:\:- & 3.05 & 5.53 & \:\:\:\:- \\
&mGP, only B62 & 0.72 & -0.53 & \:\:\:\:- & 0.08 & 0.07 & \:\:\:\:- & 0.30 & -0.00 & \:\:\:\:- &1.53 & 2.01 & \:\:\:\:- & 6.89 & 6.77 & \:\:\:\:- & 3.14 & 5.41 & \:\:\:\:- \\
&Rosetta scaled & 0.58 & 0.43 & \:\:\:\:- & 0.78 & 0.86 & \:\:\:\:- & 0.42 & 0.51 & \:\:\:\:- &1.57 & 1.76 & \:\:\:\:- & 7.08 & 7.24 & \:\:\:\:- & 3.23 & 5.74 & \:\:\:\:- \\ \cline{2-20}
&Rosetta & \multicolumn{3}{c|}{0.60} &\multicolumn{3}{c|}{0.77} & \multicolumn{3}{c||}{0.39} & \multicolumn{3}{c|}{1.58} & \multicolumn{3}{c|}{6.90} & \multicolumn{3}{c}{3.16} \\
	&mCSM	    &\multicolumn{3}{c|}{0.77} &\multicolumn{3}{c|}{-}	&\multicolumn{3}{c||}{-}	&\multicolumn{3}{c|}{1.36} &\multicolumn{3}{c|}{-}	&\multicolumn{3}{c}{-}	\\
	&PoPMuSiC	&\multicolumn{3}{c|}{0.65} &\multicolumn{3}{c|}{-}	&\multicolumn{3}{c||}{-}	&\multicolumn{3}{c|}{1.54} &\multicolumn{3}{c|}{-}	&\multicolumn{3}{c}{-}	\\
\hline
\multirow{6}{*}{1BPI}	&mGPfusion & 0.68 & 0.67 & 0.51 & -0.39 & -0.02 & -0.06 & 0.85 & -0.13 & 0.12 & 1.27 & 1.28 & 1.83 & 2.17 & 7.66 & 7.69 & 1.40 & 3.64 & 3.11 \\
&mGPfusion, only B62 & 0.64 & 0.65 & 0.51 & -0.53 & 0.33 & 0.21 & 0.81 & 0.05 & 0.12 &1.32 & 1.32 & 1.82 & 2.86 & 7.07 & 7.70 & 1.57 & 3.40 & 3.11 \\
&mGP & 0.56 & 0.57 & \:\:\:\:- & -0.29 & 0.52 & \:\:\:\:- & 0.83 & 0.84 & \:\:\:\:- &1.43 & 1.43 & \:\:\:\:- & 1.72 & 4.19 & \:\:\:\:- & 1.47 & 2.28 & \:\:\:\:- \\
&mGP, only B62 & 0.43 & 0.58 & \:\:\:\:- & -0.55 & 0.46 & \:\:\:\:- & 0.76 & 0.70 & \:\:\:\:- &1.68 & 1.73 & \:\:\:\:- & 2.52 & 6.58 & \:\:\:\:- & 1.80 & 3.34 & \:\:\:\:- \\
&Rosetta scaled & 0.52 & 0.52 & \:\:\:\:- & 0.23 & -0.00 & \:\:\:\:- & 0.08 & -0.08 & \:\:\:\:- &1.52 & 1.52 & \:\:\:\:- & 7.43 & 7.43 & \:\:\:\:- & 2.88 & 3.62 & \:\:\:\:- \\ \cline{2-20}
&Rosetta & \multicolumn{3}{c|}{0.51} &\multicolumn{3}{c|}{-0.00} & \multicolumn{3}{c||}{0.13} & \multicolumn{3}{c|}{1.80} & \multicolumn{3}{c|}{7.68} & \multicolumn{3}{c}{3.09} \\
	&mCSM	    &\multicolumn{3}{c|}{0.71} &\multicolumn{3}{c|}{-}	&\multicolumn{3}{c||}{-}	&\multicolumn{3}{c|}{1.26} &\multicolumn{3}{c|}{-}	&\multicolumn{3}{c}{-}	\\
	&PoPMuSiC	&\multicolumn{3}{c|}{0.72} &\multicolumn{3}{c|}{-}	&\multicolumn{3}{c||}{-}	&\multicolumn{3}{c|}{1.31} &\multicolumn{3}{c|}{-}	&\multicolumn{3}{c}{-}	\\
\hline
\multirow{6}{*}{\bf{total}} &mGPfusion & 0.81 & 0.70 & 0.56 & 0.88 & 0.61 & 0.49& 0.83 & 0.64 & 0.52 & 1.07 & 1.26 & 1.61 & 1.33 & 2.45 & 2.53 & 1.13 & 1.87 & 1.84 \\
&mGPfusion, only B62 & 0.79 & 0.69 & 0.56 & 0.86 & 0.64 & 0.50 & 0.82 & 0.66 & 0.52 & 1.11 & 1.30 & 1.62 & 1.43 & 2.40 & 2.50 & 1.18 & 1.85 & 1.84 \\
&mGP & 0.81 & 0.51 & \:\:\:\:- & 0.86 & 0.52 & \:\:\:\:- & 0.83 & 0.50 & \:\:\:\:- & 1.04 & 1.54 & \:\:\:\:- & 1.44 & 2.65 & \:\:\:\:- & 1.14 & 2.09 & \:\:\:\:-\\
&mGP, only B62 & 0.76 & 0.34 & \:\:\:\:- & 0.86 & 0.55 & \:\:\:\:- & 0.80 & 0.49 & \:\:\:\:- & 1.26 & 1.95 & \:\:\:\:- & 1.45 & 2.56 & \:\:\:\:- & 1.30 & 2.23 & \:\:\:\:- \\
&Rosetta scaled & 0.65 & 0.63 & \:\:\:\:- & 0.51 & 0.39 & \:\:\:\:- & 0.60 & 0.48 & \:\:\:\:- & 1.35 & 1.38 & \:\:\:\:- & 2.49 & 2.99 & \:\:\:\:- & 1.66 & 2.22 & \:\:\:\:- \\ \cline{2-20}
&Rosetta & \multicolumn{3}{c|}{0.55} &\multicolumn{3}{c|}{0.40} & \multicolumn{3}{c||}{0.49} & \multicolumn{3}{c|}{1.63} & \multicolumn{3}{c|}{2.74} & \multicolumn{3}{c}{1.92} \\
    &mCSM     & \multicolumn{3}{c|}{0.61} & \multicolumn{3}{c|}{-}  & \multicolumn{3}{c||}{-} & \multicolumn{3}{c|}{1.40} & \multicolumn{3}{c|}{-}  & \multicolumn{3}{c}{-} \\
    &PoPMuSiC & \multicolumn{3}{c|}{0.64} & \multicolumn{3}{c|}{-}  & \multicolumn{3}{c||}{-} & \multicolumn{3}{c|}{1.37} & \multicolumn{3}{c|}{-}  & \multicolumn{3}{c}{-} \\
\bottomrule
\end{tabular}}}{ \footnotesize $^*$~dataset for 1PIN contained no multiple mutations.}
\end{table}}

\clearpage

{\begin{table*}[t]
\caption{Comparison of different methods on the $15$ protein dataset with respect to $\rho$ and $\rmse$ after removing 10\% of predictions with largest errors. Mutation, position, and protein are referred to as mut., pos., and prot., respectively. Off-the-shelf implementations of Rosetta, mCSM and PoPMuSiC are used directly without cross-validation.} \label{tab:comparison10}
{\resizebox{\linewidth}{!}{
\begin{tabular}{@{} l | rcl | rcl | rcl || rcl | rcl | rcl @{}}\toprule 
\multirow{3}{*}{\shortstack{Results after 10 \% \\ outlier removal}}& \multicolumn{9}{c||}{Correlation $\rho$} & \multicolumn{9}{c}{$\rmse$}\\
& \multicolumn{3}{c|}{Point mutations} & \multicolumn{3}{c|}{Multiple mutations} & \multicolumn{3}{c||}{All mutations} & \multicolumn{3}{c|}{Point mutations} & \multicolumn{3}{c|}{Multiple mutations} & \multicolumn{3}{c}{All mutations} \\
\cline{2-19}
& \multicolumn{3}{c|}{ cross-validation level }& \multicolumn{3}{c|}{ cross-validation level }& \multicolumn{3}{c||}{ cross-validation level }& \multicolumn{3}{c|}{ cross-validation level }& \multicolumn{3}{c|}{ cross-validation level }& \multicolumn{3}{c}{ cross-validation level }\\
Method & \:\,  mut. & pos. & \multicolumn{1}{l|}{ prot.} & \:\,  mut. & pos. & \multicolumn{1}{l|}{ prot.} & \:\,  mut. & pos. & prot. & \:\,  mut. & pos. & \multicolumn{1}{l|}{ prot.} & \:\,  mut. &  pos. & \multicolumn{1}{l|}{ prot.} & \:\,  mut. & pos. & prot.\\
\hline
mGPfusion & 0.87 & 0.77 & 0.75 & 0.97 & 0.84 & 0.63& 0.92 & 0.81 & 0.71 & 0.69 & 0.85 & 0.93 & 0.62 & 1.55 & 1.90 & 0.67 & 1.15 & 1.13 \\
mGPfusion, only B62 & 0.86 & 0.73 & 0.75 & 0.96 & 0.84 & 0.65 & 0.91 & 0.80 & 0.72 & 0.69 & 0.85 & 0.93 & 0.64 & 1.43 & 1.85 & 0.67 & 1.10 & 1.12 \\
mGP & 0.89 & 0.57 & \:\:\:\:- & 0.98 & 0.76 & \:\:\:\:- & 0.93 & 0.74 & \:\:\:\:- & 0.65 & 1.01 & \:\:\:\:- & 0.49 & 1.66 & \:\:\:\:- & 0.62 & 1.23 & \:\:\:\:-\\
mGP, only B62 & 0.86 & 0.27 & \:\:\:\:- & 0.97 & 0.76 & \:\:\:\:- & 0.91 & 0.66 & \:\:\:\:- & 0.73 & 1.30 & \:\:\:\:- & 0.54 & 1.48 & \:\:\:\:- & 0.69 & 1.37 & \:\:\:\:- \\
Rosetta scaled & 0.80 & 0.78 & \:\:\:\:- & 0.73 & 0.71 & \:\:\:\:- & 0.78 & 0.75 & \:\:\:\:- & 0.84 & 0.87 & \:\:\:\:- & 1.67 & 1.99 & \:\:\:\:- & 1.00 & 1.31 & \:\:\:\:- \\\midrule
\multicolumn{19}{l}{Off-the-shelf implementations with no cross-validation}\\\midrule
Rosetta & \multicolumn{3}{c|}{0.75} &\multicolumn{3}{c|}{0.67} & \multicolumn{3}{c||}{0.73} & \multicolumn{3}{c|}{0.94} & \multicolumn{3}{c|}{1.85} & \multicolumn{3}{c}{1.11}\\
mCSM     & \multicolumn{3}{c|}{0.71} & \multicolumn{3}{c|}{-}  & \multicolumn{3}{c||}{-} & \multicolumn{3}{c|}{0.89} & \multicolumn{3}{c|}{-}  & \multicolumn{3}{c}{-} \\
PoPMuSiC & \multicolumn{3}{c|}{0.73} & \multicolumn{3}{c|}{-}  & \multicolumn{3}{c||}{-} & \multicolumn{3}{c|}{0.86} & \multicolumn{3}{c|}{-}  & \multicolumn{3}{c}{-} \\\bottomrule
\end{tabular}}}{}
\end{table*}}

\bigskip
\begin{figure}[H]
\centerline{\includegraphics[width=\linewidth]{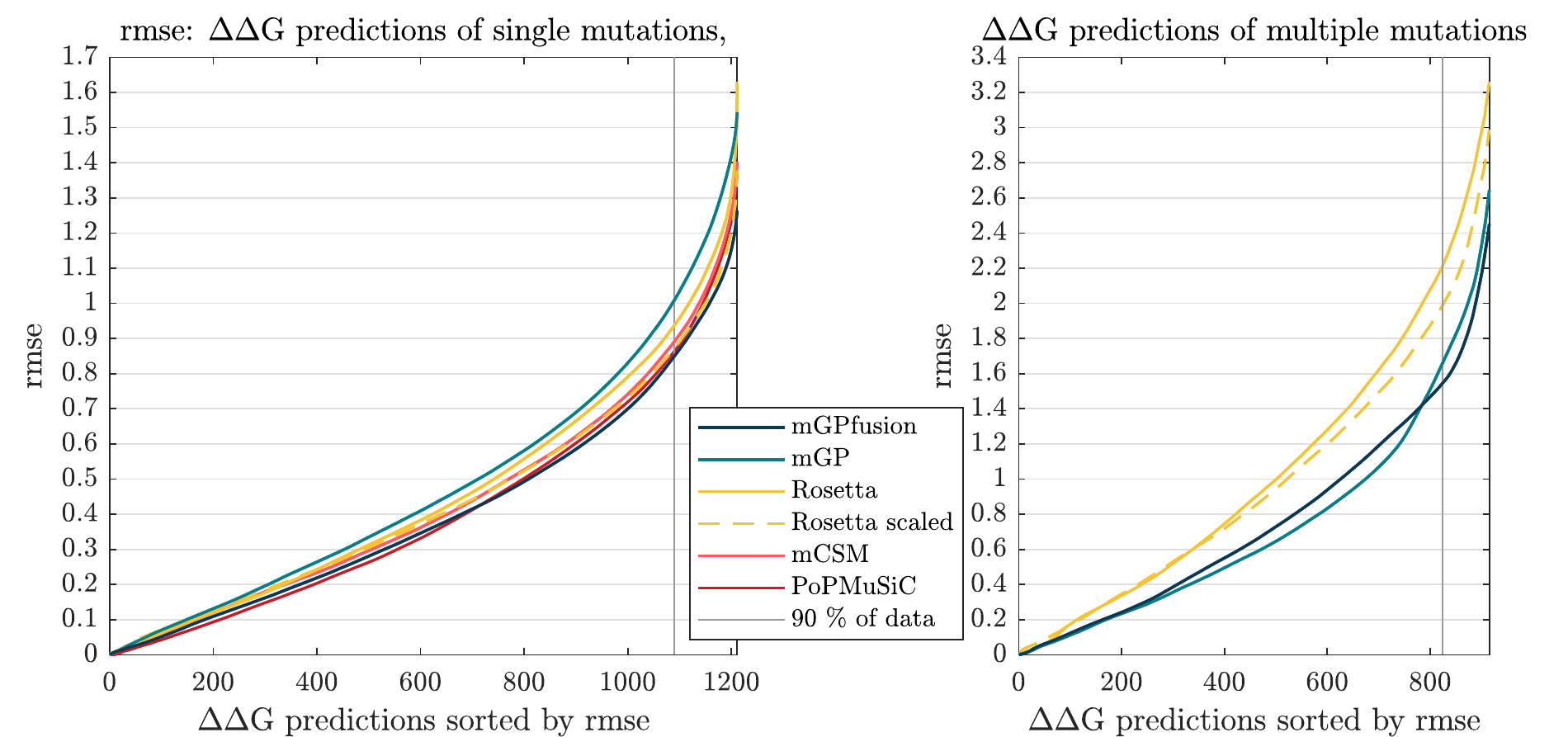}}
\caption{$\rmse$ with different amount of predictions, when predictions are sorted by the error. Position level cross-validation was used for mGPfusion, mGP and Rosetta scaled.}\label{fig:rmse}
\end{figure}
\vfill

\begin{figure}[!htb]
\centerline{\includegraphics[width=\linewidth]{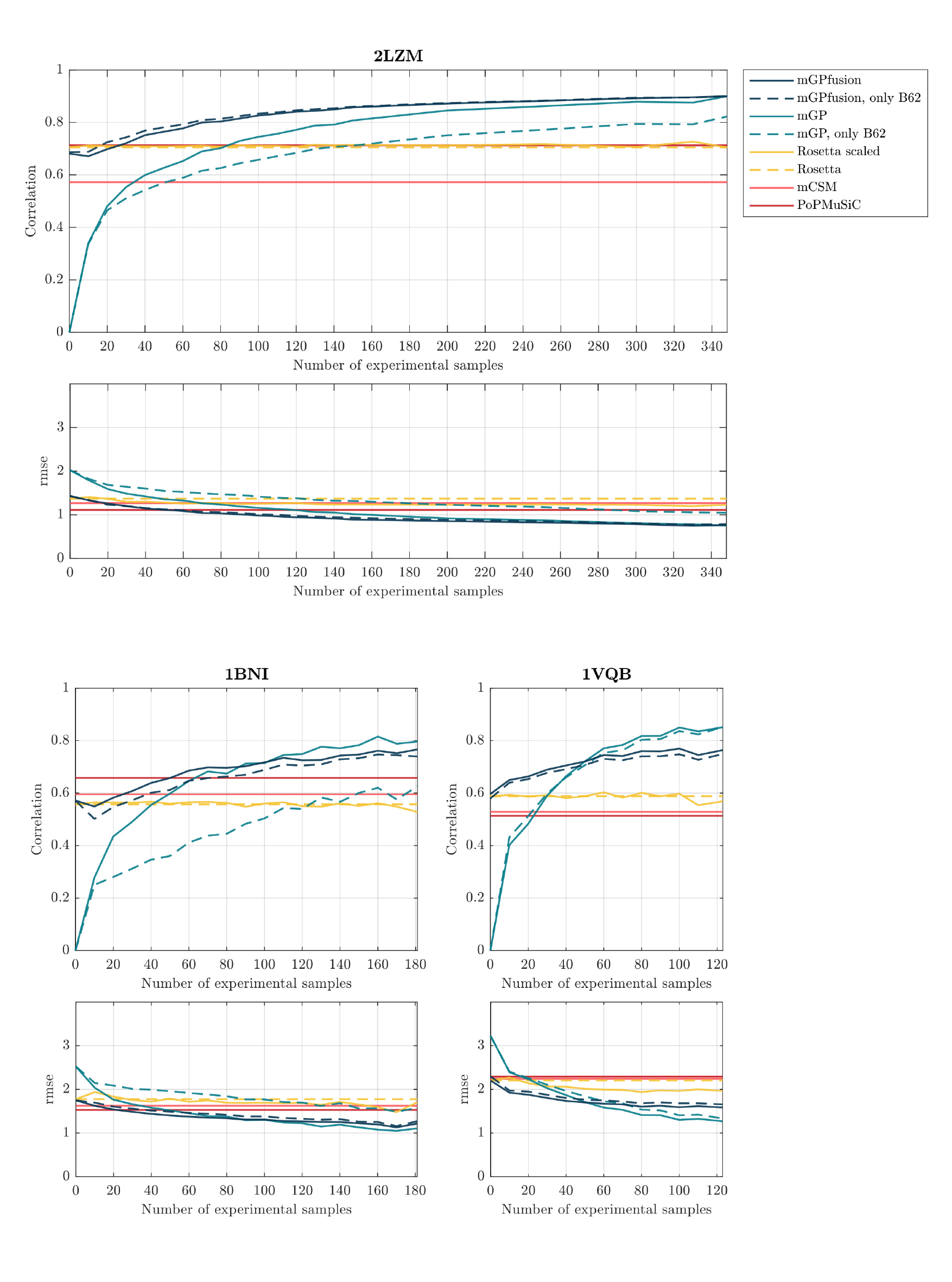}}
\caption{Learning curves.}\label{fig:curves1}
\end{figure}
\begin{figure}[!htb]
\centerline{\includegraphics[width=\linewidth]{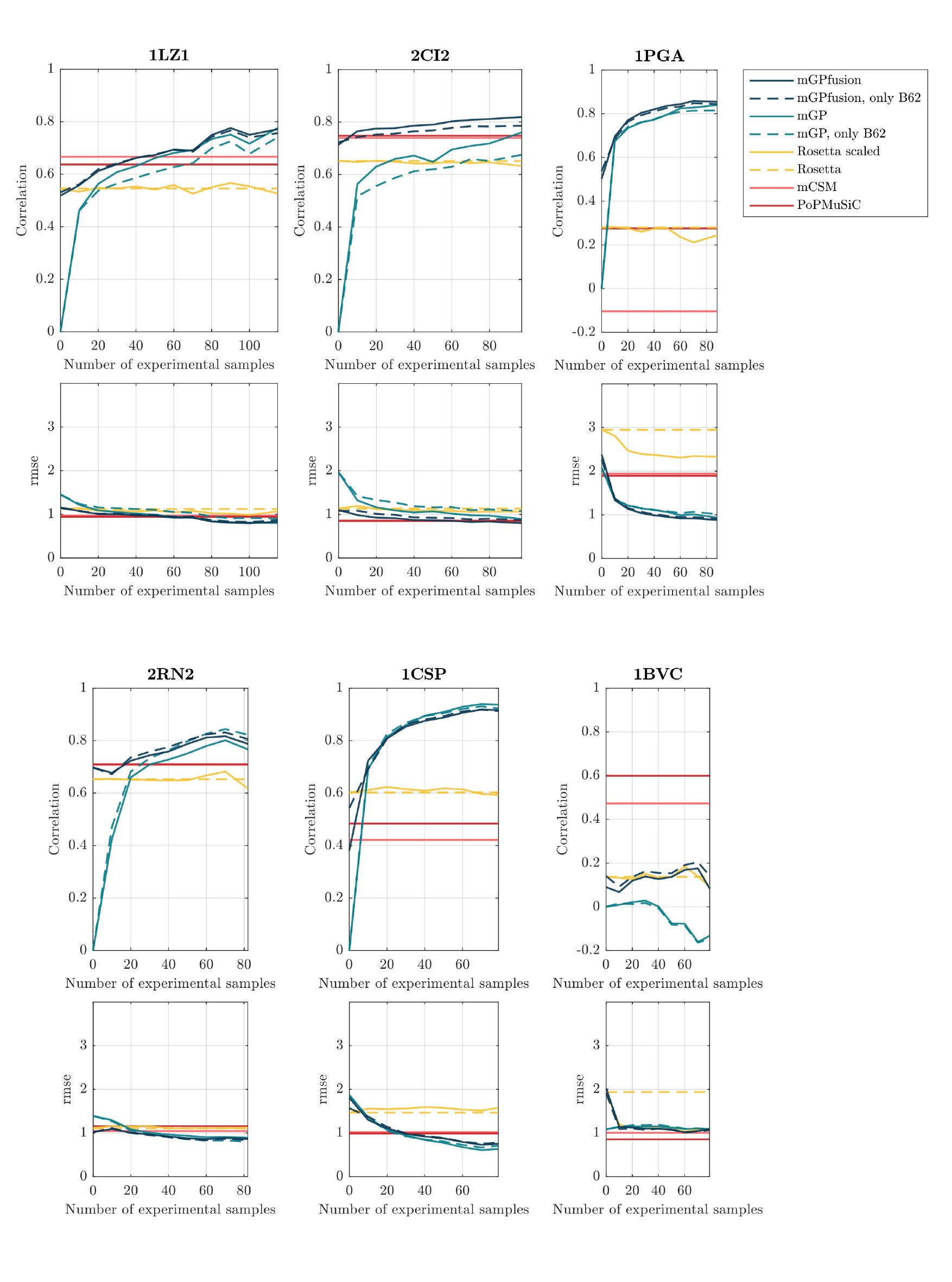}}
\caption{Learning curves.}\label{fig:curves2}
\end{figure}
\begin{figure}[!htb]
\centerline{\includegraphics[width=\linewidth]{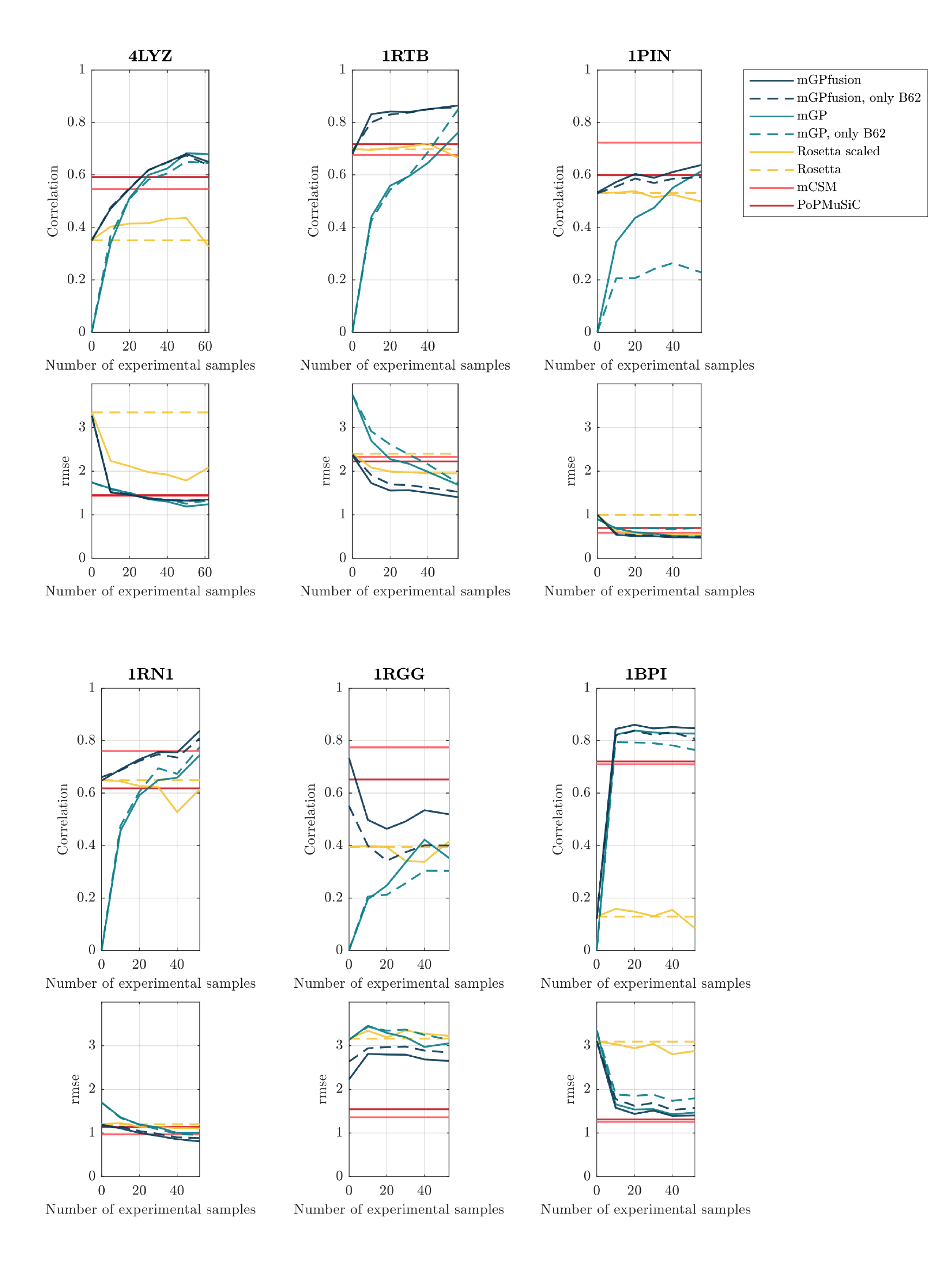}}
\caption{Learning curves.}\label{fig:curves3}
\end{figure}

\begin{figure}[!htb]
\centerline{\includegraphics[width=\linewidth]{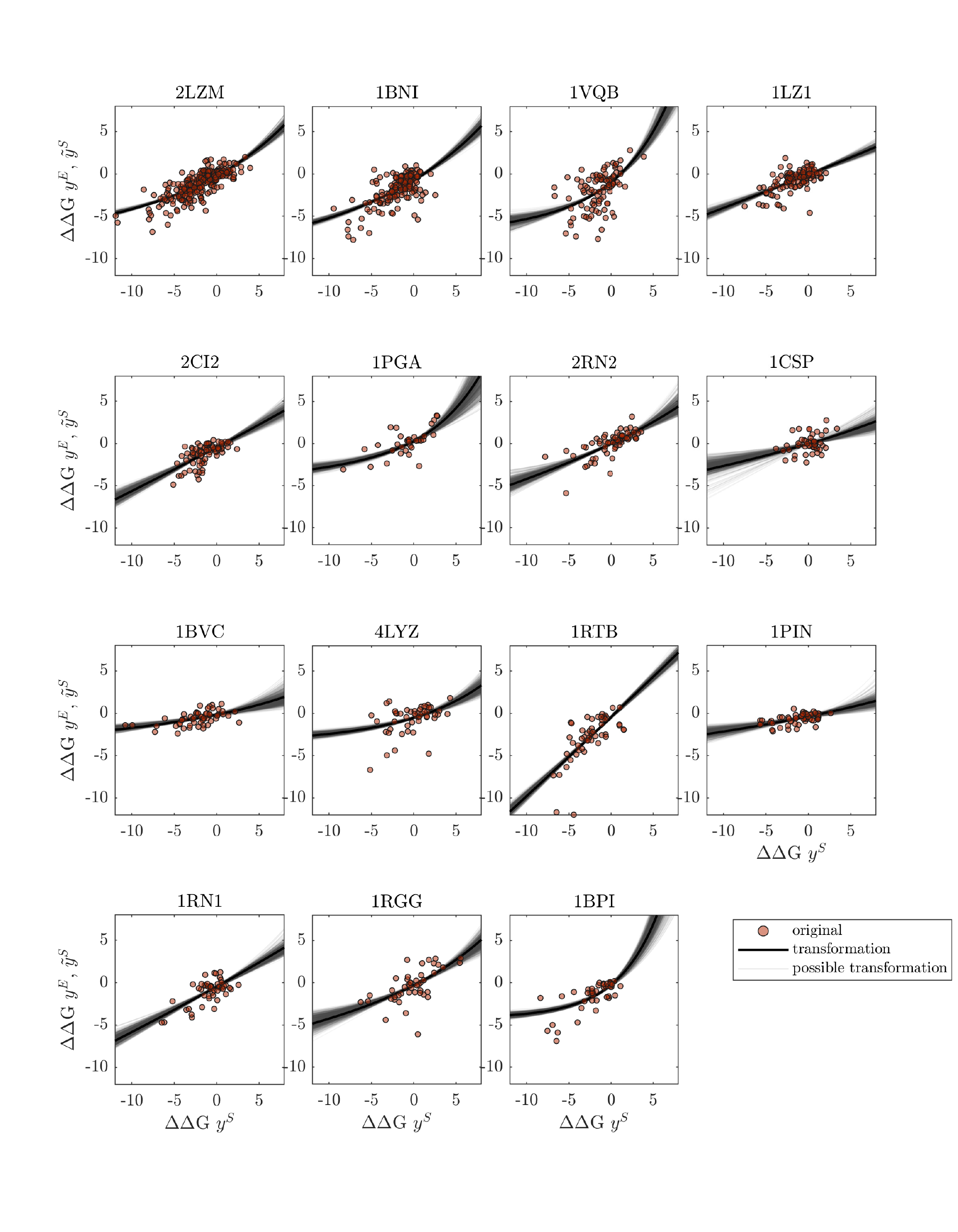}}
\caption{Transformations for all 15 proteins presented in Table~1. The red circles mark the simulated $\DDG$-values $y^S$ with respect to the experimental measured $\DDG$-values $y^E$. Thin black lines show possible transformations for $y^S$, whereas the thick black line shows the selected transformation from $y^S$ to $\tilde{y}^S$.}\label{fig:transformation3}
\end{figure}

\end{document}